\documentclass{article}

\PassOptionsToPackage{numbers, compress}{natbib}

\usepackage[final,main]{neurips_2026}

\usepackage[utf8]{inputenc} 
\usepackage[T1]{fontenc}    
\usepackage{hyperref}       
\usepackage{url}            
\usepackage{booktabs}       
\usepackage{amsfonts}       
\usepackage{nicefrac}       
\usepackage{microtype}      
\usepackage{xcolor}         
\usepackage{amsmath,amssymb,mathrsfs,bm,latexsym,graphicx}
\usepackage{multirow}
\usepackage{paralist}
\usepackage{setspace,color,verbatim}
\usepackage{algorithm}
\usepackage{algpseudocode}
\usepackage{graphicx}
\usepackage{amsthm}
\newtheorem{theorem}{Theorem}

\newtheorem{assumption}{Assumption}
\newtheorem{lemma}{Lemma}
\usepackage{dsfont}

\usepackage{tabularx}
\usepackage{array}
\usepackage{hyperref} 
\usepackage{cleveref} 
\usepackage{bbm} 
\usepackage{makecell} 
\usepackage{pifont}
\newcolumntype{Y}{>{\centering\arraybackslash}X} 
\newcommand{\perturb}{\ensuremath{^\clubsuit}}  
\usepackage{microtype}
\usepackage{graphicx}
\usepackage{subcaption}
\usepackage{booktabs} 
\usepackage{makecell}
\usepackage{tabularx}
\usepackage{wrapfig}

\crefname{assumption}{assumption}{assumptions}   
\Crefname{assumption}{Assumption}{Assumptions}   

\title{Alignment Imprint: Zero-Shot AI-Generated Text Detection via Provable Preference Discrepancy}

\author{Junxi Wu$^{1,2}$\thanks{Equal contribution.}~, \quad
Kailin Huang$^{3}$\footnote[1]{}~, \quad
Dongjian Hu$^{1}$\footnote[1]{}~, \quad
Bin Chen$^{4}$\thanks{Corresponding author.}~, \\
\textbf{Hao Wu$^{2,5}$, \quad
Shu-Tao Xia$^{2}$, \quad
Changliang Zou$^{1}$}\\
$^{1}$ Nankai University \quad $^{2}$ Tsinghua University \quad 
$^{3}$ Wuhan University\\
$^{4}$ Harbin Institute of Technology, Shenzhen \quad
$^{5}$ Shannon InfoTech\\
\texttt{wujunxi@mail.nankai.edu.cn}\\
}

\begin{document}

\maketitle

\begin{abstract}

Detecting AI-generated text is an important but challenging problem. Existing likelihood-based detection methods are often sensitive to content complexity and may exhibit unstable performance.
In this paper, our key insight is that modern Large Language Models (LLMs) undergo alignment (including fine-tuning and preference tuning), leaving a measurable distributional imprint. 
We theoretically derive this imprint by abstracting the alignment process as a sequence of constrained optimization steps, showing that the log-likelihood ratio can naturally decompose into implicit instructional biases and preference rewards. We refer to this quantity as the \textit{Alignment Imprint}.
Furthermore, to mitigate the instability in high-entropy regions, we introduce \textit{Log-likelihood Alignment Preference Discrepancy (LAPD)}, a standardized information-weighted statistic based on alignment imprint. 
We provide statistical guarantee that alignment-based statistics dominate Fast-DetectGPT in performance. We also theoretically show that LAPD strictly improves the unweighted alignment scores when the aligned and base models are close in distribution.
Extensive experiments show that LAPD achieves an improvement $45.82\%$ relative to the strongest existing baselines, yielding large and consistent gains across all settings. Our code is available at \url{https://github.com/creator-xi/LAPD}.
\end{abstract}

\section{Introduction}

\begin{figure*}
    \centering
    \includegraphics[width=\textwidth]{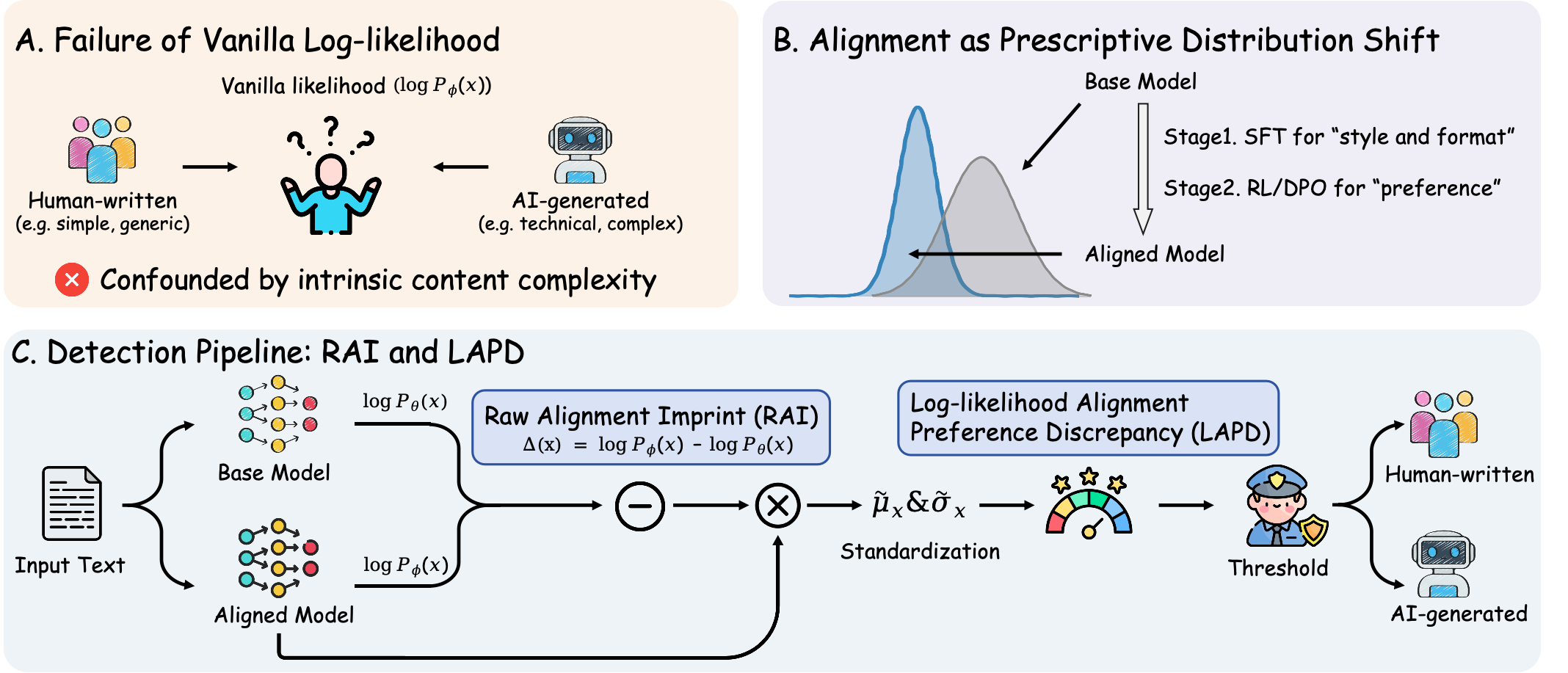}
    \caption{Overview. \textbf{(A)} Vanilla log-likelihood is confounded by intrinsic content complexity (e.g. simple human-written text may receive higher likelihood than complex AI-generated text), undermining reliable detection. \textbf{(B)} LLM alignment pipelines induce a prescriptive distribution shift, yielding a systematic shift between base and aligned models. \textbf{(C)} Detection pipeline. 
    }
    \label{LAPD_framework}
    \vspace{-1em}
\end{figure*}

The rapid advancement of Large Language Models (LLMs) has significantly expanded the capabilities of automated text generation. 
However, these capabilities also introduce profound risks, including the spread of disinformation, academic dishonesty, and malicious text attacks \citep{opdahl2023trustworthy}. 
Modern LLMs produce fluent and coherent content, blurring the statistical boundary with human-written text.
Consequently, developing more reliable mechanisms to distinguish AI-generated text has become a critical requirement for mitigating misuse.

A popular line of recent work frames detection as a likelihood-based problem, leveraging the observation that AI-generated text tends to receive higher probability under the generative model. However, such approaches remain fundamentally confounded by intrinsic content complexity. A simple human-written sentence may receive a higher likelihood than a long technical paragraph generated by an AI system \citep{wu2025moses}. As a result, likelihood score alone often collapses in the real world.

In practice, users almost exclusively interact with \textit{aligned models}, rather than raw pre-trained \textit{base models}, which are not designed to reliably follow instructions. Modern LLMs (e.g., ChatGPT \citep{achiam2023gpt}, Gemini \citep{comanici2025gemini}, and DeepSeek \citep{guo2025deepseek}) are typically first trained on large-scale unsupervised corpora and then transformed through a multi-stage alignment process. This process commonly consists of supervised fine-tuning (SFT), followed by preference tuning methods such as RLHF \citep{ouyang2022rlhf} or DPO \citep{rafailov2023dpo}. SFT adapts the base model to instruction–response data drawn from curated datasets, while preference tuning further adjusts the output distribution to satisfy explicit reward objectives under KL regularization. Together, these stages systematically modify the model’s output distribution by introducing prescriptive constraints that reflect safety norms and human preferences, echoing the conclusions drawn by \citet{sivaprasad2025theory}.

Our key insight is that the alignment process leaves a distributional imprint that is statistically exploitable to distinguish. While human language arises from diverse communicative intents, AI-generated text is optimized to satisfy explicit and implicit alignment objectives. As a consequence, aligned models systematically reallocate the probability mass toward a narrower subset of ``preferred'' responses, inducing entropy contraction and mode concentration. \citet{kirk2024understanding} have observed this effect and reported reduced lexical and semantic diversity following alignment.

Given a pair of a base model and an aligned model, we characterize the distributional shift induced by alignment and derive that the log-likelihood ratio between them can naturally decompose into implicit instructional biases introduced during SFT and explicit rewards imposed during alignment. We refer to this quantity as the \textit{Alignment Imprint}. Intuitively, it measures how much additional probability a text receives solely due to alignment.

However, directly using the raw log-likelihood ratio as a detection statistic remains suboptimal. In high-entropy regions, small aleatoric variations can induce large fluctuations in log space, masking the alignment signal. To address this issue, we further introduce \textit{\textbf{L}og-likelihood \textbf{A}lignment \textbf{P}reference \textbf{D}iscrepancy (LAPD)}, a standardized information-weighted statistic based on alignment imprint.
By weighting the alignment imprint according to predictability, LAPD amplifies the imprint and achieves much better performance.

Furthermore, we provide theoretical guarantees for the superiority of the alignment imprint and LAPD. Under mild and interpretable assumptions, we prove that alignment-based statistics dominate Fast-DetectGPT \citep{bao2024fast} in terms of true negative rate at any fixed false negative rate. 
Moreover, we show that LAPD strictly improves upon unweighted alignment scores in regimes where the aligned model and base model are close in distribution.

Our contributions are summarized as follows: 
\begin{itemize} 
\vspace{-1em}
\item We derive a principled characterization by abstracting alignment pipelines as two constrained optimization steps. We show that the log-likelihood ratio between aligned model and its base model can naturally decompose into implicit instructional biases and preference rewards. This derivation formalizes the alignment imprint as a measurable consequence, without requiring access to reward models or training data.
\vspace{-0.5em}
\item Building on alignment imprint, we introduce LAPD, a standardized statistic that combines alignment-induced likelihood ratios with token-level self-information. This enhancement yields a stable and effective detection score that consistently outperforms existing detectors across LLMs, domains, and text lengths.
\vspace{-0.5em}
\item We provide two theoretical guarantees on detection performance. Under mild and interpretable assumptions, we prove that the alignment-based statistics dominate Fast-DetectGPT in terms of true negative rate at any fixed false negative rate. Moreover, we also theoretically show that LAPD strictly improves the unweighted alignment scores when the aligned and base models are close in distribution.
\vspace{-0.5em}
\item We have conducted extensive experiments on four comprehensive benchmarks. LAPD yields large and consistent gains across all benchmarks, achieving an average relative improvement of $56.99\%$ over Fast-DetectGPT and $45.82\%$ over the strongest baselines. Additional six robustness and ablation studies confirm that the improvements are stable in complex practical scenarios, with each component consistently contributing to effectiveness. In particular, LAPD improves $76.81\%$ under an extremely $0.5\%$ Type-I error constraint.
\vspace{-1em}
\end{itemize}

\begin{figure*}[t]
  \centering
  \begin{minipage}[c]{0.02\textwidth} 
    \centering
    \rotatebox{90}{\footnotesize Frequency} 
  \end{minipage}
  \hfill
  \begin{minipage}[c]{0.97\textwidth} 
    \centering
    
    \begin{subfigure}[b]{0.24\linewidth} 
      \centering
      \includegraphics[width=\linewidth]{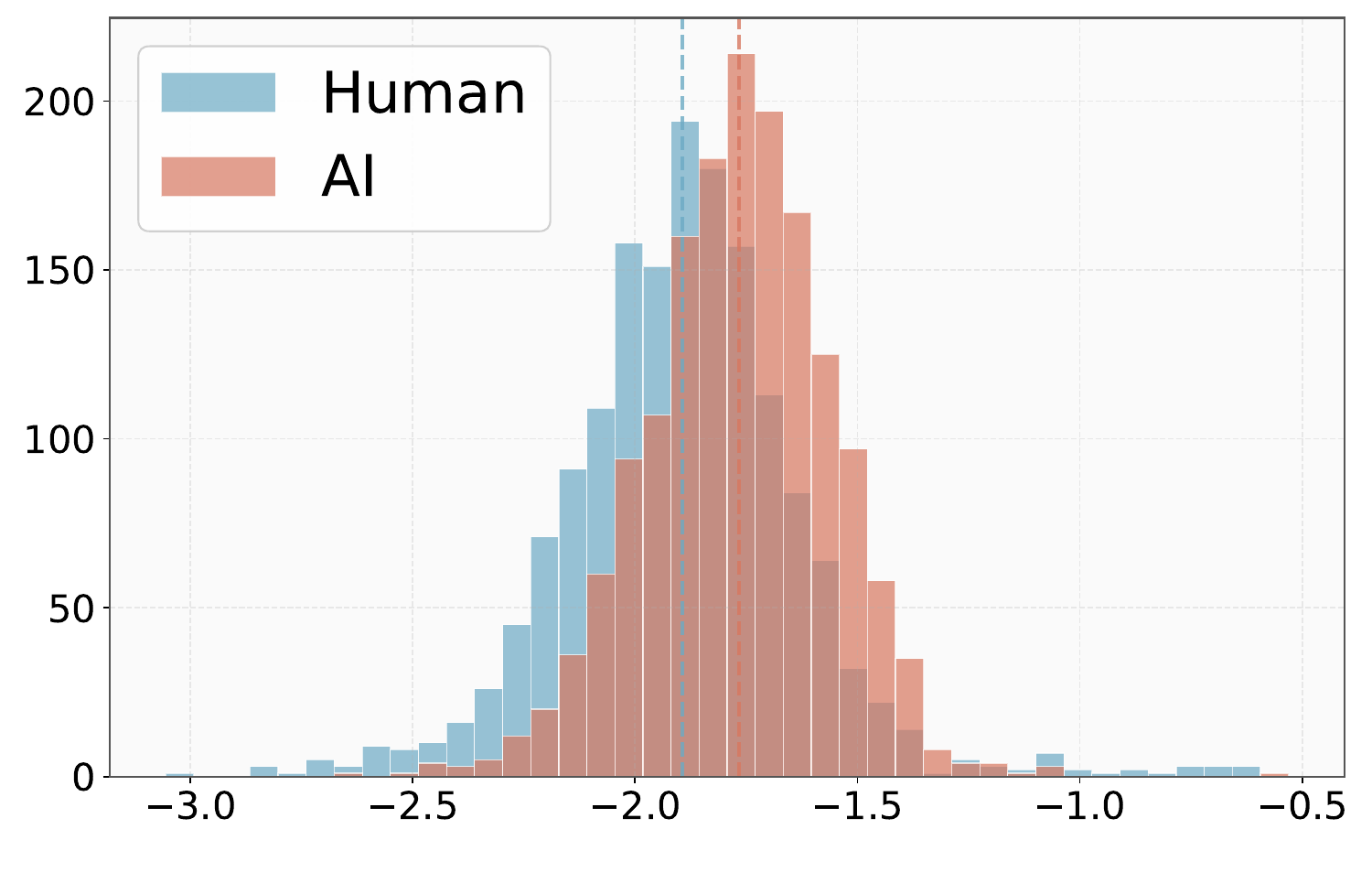} 
      \caption{Base Model}
      \label{fig:dist_base}
    \end{subfigure}
    \hfill 
    \begin{subfigure}[b]{0.24\linewidth}
      \centering
      \includegraphics[width=\linewidth]{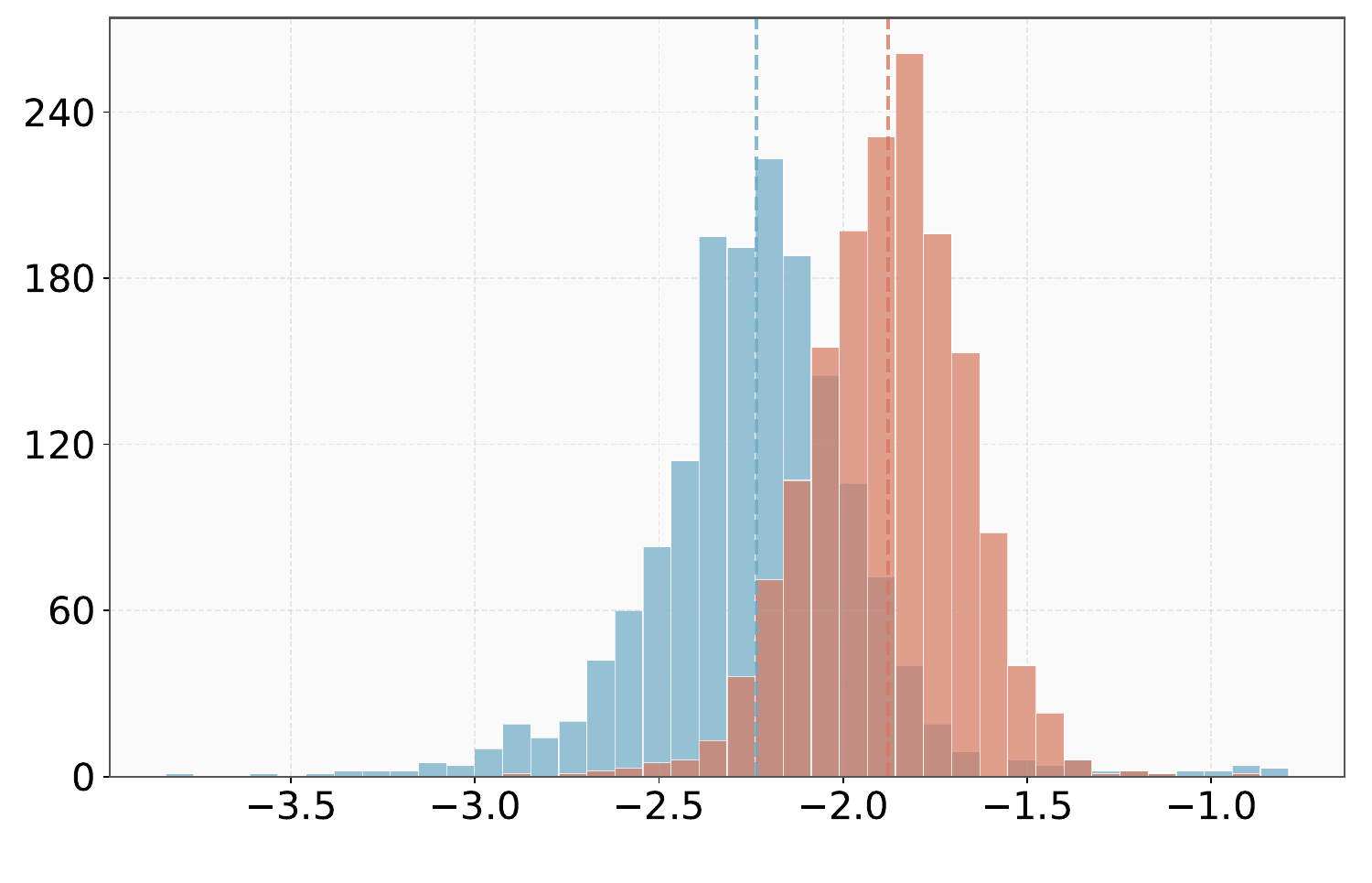}
      \caption{Aligned Model}
      \label{fig:dist_aligned}
    \end{subfigure}
    \hfill
    \begin{subfigure}[b]{0.24\linewidth}
      \centering
      \includegraphics[width=\linewidth]{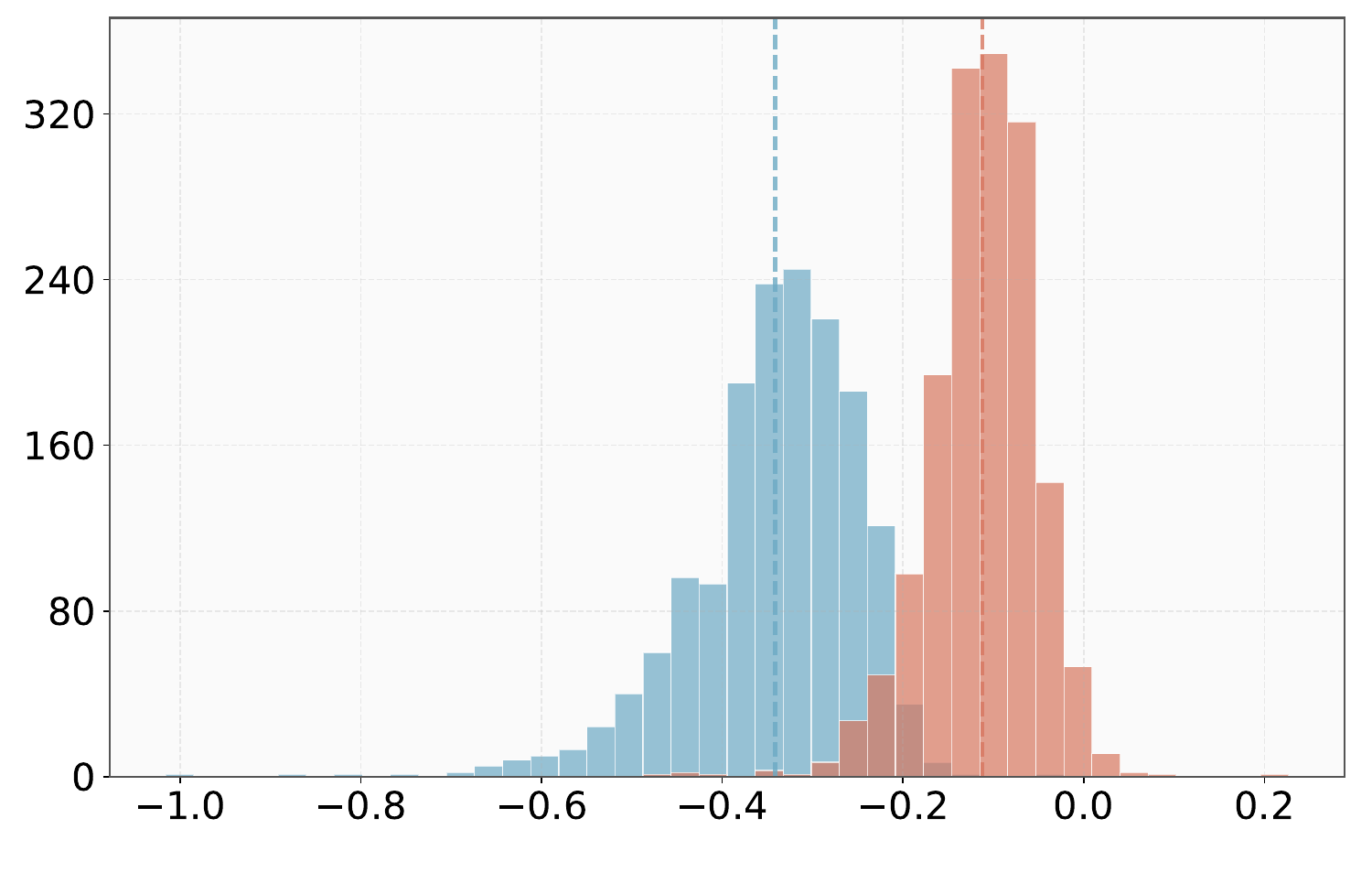}
      \caption{RAI}
      \label{fig:dist_rai}
    \end{subfigure}
    \hfill
    \begin{subfigure}[b]{0.24\linewidth}
      \centering
      \includegraphics[width=\linewidth]{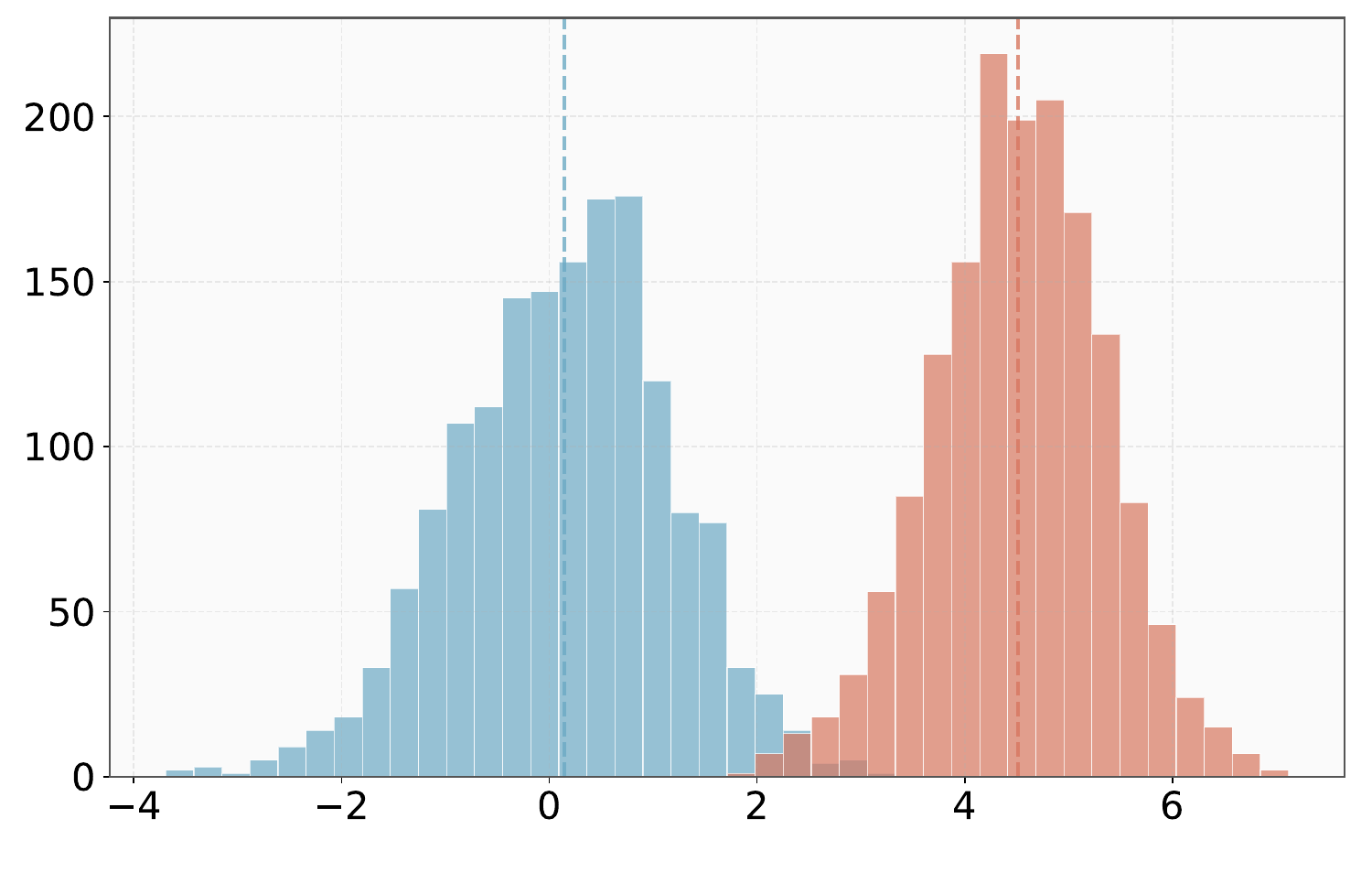}
      \caption{LAPD}
      \label{fig:dist_lapd}
    \end{subfigure}
  \end{minipage}
  \caption{Distributions of human-written texts and AI-generated texts. Dashed vertical lines represent the mean value of each distribution. (a) Average log-likelihood under the base model; (b) average log-likelihood under the aligned model; (c) raw alignment imprint
  ; and (d) LAPD score.}
  \vspace{-1em}
  \label{fig:dist}
\end{figure*}

\section{Related Work}
\subsection{LLM-Generated Text Detection}

Existing methods can be categorized into training-based and training-free approaches.
Training-based methods \citep{solaiman2019release, verma2024ghostbuster, hu2023radar} involve training a classifier on collected datasets for detection. There are also methods like ImBD \citep{chen2025imbd} that fine-tunes a proxy LLM to favor machine preferences for more sensitive detection.
However, these supervised models often suffer from the overfitting problem \citep{chakraborty2024position}.

Training-free methods leverage the probabilistic features of the text without additional training, exhibiting superior generalization.
Early approaches used simple metrics such as Entropy \citep{ippolito2020entropy}, Log-Likelihood \citep{solaiman2019release}, and LogRank \citep{gehrmann2019gltr}.
More recently, DetectGPT \citep{mitchell2023detectgpt} introduced a perturbation paradigm, utilizing separate perturbing and scoring models for detection. Fast-DetectGPT \citep{bao2024fast} further enhances efficiency by conditional sampling, while Lastde \citep{xu2025lastde} incorporates temporal dynamics. Additionally, Binoculars \citep{hans2024binoculars} contrasts perplexity with cross-perplexity, and DNA-DetectLLM \citep{zhu2025dna} employs a repair score based on a mutation-repair mechanism. 

Most closely related to our work are that use difference between base and aligned LLMs, including ReMoDetect \cite{lee2024remodetect}, IRM \cite{liu2025zero}, and dual-network \cite{chen2026zero}. 
While these works explore the alignment gap between model pairs, our work provides the first complete theoretical derivation of the alignment imprint across the full SFT and preference tuning pipeline, introduces the noise-robust information-weighted LAPD statistic and offers formal provable performance guarantees, in contrast to ReMoDetect’s training-dependent pipeline and the limited theoretical framing of IRM and dual-network method.

\subsection{LLM Alignment}

The pipeline of modern LLMs training usually progresses through three steps: Pre-training, Supervised Fine-Tuning (SFT), and preference tuning (e.g., RLHF or DPO) \citep{zhao2023llm-survey}.
Pre-training establishes the fundamental probability distribution over massive corpora data. Subsequently, SFT adapts the base model to follow user instructions by maximizing the likelihood estimation of high-quality instructions. This process can be conceptualized as reweighting the pretrained distribution to favor coherent and instruction-following sequences \citep{zhou2023lima}.

Preference tuning is used to align models with human values such as safety and helpfulness. 
The traditional approach, RLHF \citep{ouyang2022rlhf}, optimizes a policy against a learned reward model with algorithms like PPO \citep{schulman2017ppo}. It often constrains the generation diversity to ensure safety \citep{kirk2024understanding}.
Recently, DPO \citep{rafailov2023dpo} has emerged as an alternative, which analytically solves the RLHF objective and optimizes the policy directly from preference data, implicitly modeling the reward.
\section{Method and Theory}

In this section, we derive the \textit{Alignment Imprint} and propose our detection statistic, \textit{Log-likelihood Alignment Preference Discrepancy (LAPD)}. In particular, our objective is not to exactly replicate the training dynamics of specific algorithms, but to characterize the distributional effect induced by alignment. This abstraction allows us to derive a tractable likelihood ratio statistic, even when the true reward functions or training data are unknown.

\subsection{Deriving the Alignment Imprint}
\label{sec:align_imprint}

Let $\boldsymbol{x} = [x_0, \dots, x_T]$ denote a token sequence of length $T+1$. We denote the probability distribution of the pre-trained base model as $P_\theta(\boldsymbol{x})$\footnote{For simplicity, we denote $P_\theta(\boldsymbol{x})=\prod_{t=1}^T P_\theta(x_t|x_{<t})$, and $P_\theta(x_t)=P_\theta(x_t|x_{<t})$. The notation for $P_\phi$ follows similarly.}, representing the prior distribution learned from large-scale unsupervised corpora. Modern alignment pipelines transform this base distribution into a final aligned distribution, denoted as $P_\phi(\boldsymbol{x})$, through a two-stage process: SFT and preference tuning.

Firstly, we consider the SFT stage, where the base model is adapted to a curated dataset of instruction-response pairs. We conceptualize the construction of this dataset not as random sampling, but as a filtering process governed by a latent quality function $V(\boldsymbol{x})$, which captures implicit attributes such as instruction adherence and formatting. Applying the principle of maximum entropy under the constraint of expected quality, the theoretical SFT distribution $P_{\text{SFT}}$ is:
\begin{equation}
P_{\text{SFT}}(\boldsymbol{x}) = \frac{1}{Z_1} P_\theta(\boldsymbol{x}) \exp(V(\boldsymbol{x})),
\end{equation}

where $Z_1$ is the partition function required for normalization. This formulation implies that the probability shift in the first stage is directly proportional to the latent instructional quality $V(\boldsymbol{x})$. The full derivation is given in \Cref{appendix:sft_derivation}.

Taking the logarithm and rearranging the terms, we have:
\begin{equation}
    \log P_{\text{SFT}}(\boldsymbol{x}) - \log P_\theta(\boldsymbol{x}) = V(\boldsymbol{x}) - \log Z_1.
    \label{eq:delta sft}
\end{equation}

Subsequently, in the preference tuning stage, the model is optimized to align with human preferences. Whether optimized via RLHF or DPO, the objective can be formally defined as maximizing the expected reward $R(\boldsymbol{x})$ while minimizing the KL-divergence from the reference policy:
\begin{equation}
\max_\phi \mathbb{E}_{\boldsymbol{x} \sim P_\phi} [R(\boldsymbol{x})] - \beta D_{KL}(P_\phi || P_{\text{SFT}}),
\end{equation}

where $\beta$ is the temperature coefficient controlling the strength of the regularization, and we denote the optimal aligned policy as $P_{\phi^*}(\boldsymbol{x})$. 

The closed-form solution to this constrained optimization problem is given by:
\begin{equation}
    \log P_{\phi^*}(\boldsymbol{x}) - \log P_{\text{SFT}}(\boldsymbol{x}) = \frac{1}{\beta} R(\boldsymbol{x}) - \log Z_2,
    \label{eq:delta rl}
\end{equation}

where $Z_2 = \sum_{\boldsymbol{x}'} P_{\text{SFT}}(\boldsymbol{x}') \exp(R(\boldsymbol{x}')/\beta)$ is the normalization constant. The full derivation is given in \Cref{appendix:rl_derivation}.

By adding \cref{eq:delta sft} and \cref{eq:delta rl}, we can obtain:
\begin{equation}
\log P_{\phi^*}(\boldsymbol{x})-\log P_\theta(\boldsymbol{x})=\frac{1}{\beta} R(\boldsymbol{x}) + V(\boldsymbol{x}) - C,
\end{equation}

where $C = \log Z_1 + \log Z_2$ is a constant term and we denote the log-likelihood ratio between the final aligned model and the original base model as \textit{Raw Alignment Imprint (RAI)}:
\begin{equation}
\Delta(\boldsymbol{x}) := \log P_\phi(\boldsymbol{x}) - \log P_\theta(\boldsymbol{x}).
\end{equation}

This derivation reveals that $\Delta(\boldsymbol{x})$ quantifies the relative likelihood gain (or loss) that a text receives under the aligned model compared to the base. The term $V(\boldsymbol{x})$ accounts for the implicit ``style and format'' biases injected during SFT, while the term $R(\boldsymbol{x})$ accounts for the explicit ``value and preference'' biases injected during RLHF. 
Since LLMs are optimized to satisfy alignment objectives, AI-generated texts tend to exhibit systematically higher values of $\Delta(\boldsymbol{x})$ relative to human-written texts, even though $\Delta(\boldsymbol{x})$ is often negative due to entropy contraction induced by alignment.

\subsection{Enhancing the Alignment Imprint}

Although $\Delta(\boldsymbol{x})$ theoretically captures the divergence between aligned and base models, using it as a raw detection statistic is suboptimal due to its scale sensitivity in the log-probability space. Specifically, in high-entropy regions where model confidence is low, even minor aleatoric variations in probability can be large fluctuations in log-likelihood ratios.

We observe that AI-generated texts typically converge to a low-entropy region and positive alignment shifts. In contrast, human-written texts often traverse relatively high-entropy regions where the alignment signal is weak or negative. By coupling predictability with alignment imprint, we introduce the information-weighted score:


\begin{equation}
\mathcal{S}(\boldsymbol{x}) = \sum_{t=1}^{T} \mathcal{I}_\phi(x_t) \cdot \Delta(x_t),
\end{equation}

where $\Delta(x_t) = \log P_\phi(x_t) - \log P_\theta(x_t)$ and $\mathcal{I}_\phi(x_t) = -\log P_\phi(x_t)$ denotes the self-information of the text under the aligned model. By coupling the alignment-induced likelihood ratio with self-information, $\mathcal{S}(\boldsymbol{x})$ captures a structural property: AI-generated text is favored by alignment in confident regions, whereas human-written text tends to exhibit misalignment in regions of low confidence.

To improve robustness across different domains and sequence lengths, we standardize $\mathcal{S}(\boldsymbol{x})$ by perturbing. We define the final LAPD statistic $\mathcal{S}_{\text{LAPD}}(\boldsymbol{x})$:

\vspace{-0.3em}

\begin{equation}
\mathcal{S}_{\text{LAPD}}(\boldsymbol{x}) = \frac{\mathcal{S}(\boldsymbol{x}) - \tilde{\mu}_x}{\tilde{\sigma}_x},
\end{equation}

where $\tilde{\mu}_x$ and $\tilde{\sigma}_x$ represent the expectation and standard deviation of the score under the perturbation.
To achieve inference efficiency, we use the conditional independence sampling like \citet{bao2024fast} instead of multiple autoregressive sampling:

\vspace{-0.5em}

$$\tilde{\mu}_x = \sum_{t=1}^{T} \mathbb{E}_{x_t \sim P_\theta(\cdot \mid x_{<t})} [\mathcal{I}_\phi(x_t) \cdot \Delta(x_t)]$$
$$\tilde{\sigma}_x^2 = \sum_{t=1}^{T} \text{Var}_{x_t \sim P_\theta(\cdot \mid x_{<t})} [\mathcal{I}_\phi(x_t) \cdot \Delta(x_t)]$$

\vspace{-0.5em}

\subsection{Theoretical Guarantees}
In this section, we provide theoretical guarantees for the superiority of  LAPD. We compare the true negative rate (TNR) of LAPD with fixed FNR, versus $S_{\text{Fast}}=(\log P_{\phi}(\boldsymbol{x}) - \tilde{\mu}_x)/{\tilde{\sigma}_x},$
which was proposed by \citet{bao2024fast}. 

Note that a consistently higher TNR at fixed FNR levels implies a strictly better AUROC. With a slight abuse of notation, we denote $\tilde{\mu}_x$ and $\tilde{\sigma}_x$ as the expectation and standard deviation of $\log P_{\phi}$ under perturbation. The TNR for $S_{\text{LAPD}}$ with a threshold $\tau$ is defined as 
\begin{align*}
    \text{TNR}_{\tau}(S_{\text{LAPD}})=\mathbb{P}_{x\sim p}(S_{\text{LAPD}}(\boldsymbol{x})\leq \tau),
\end{align*}
where the probability is taken under the human-written text distribution $p$. For simplicity, we use $p_t$ and $q_t$ to denote the conditional distributions of human-written text $p(\cdot|x_{<t})$ and AI-generated text $q(\cdot|x_{<t})$, respectively.
We start by presenting the following three assumptions:
\begin{assumption}
     \label{assumption 1}
     (Indistinguishability of base model). $$\sum_{t=1}^T \mathbb{E}_{\tilde{x}_t\sim q_t}\log P_{\theta}(\tilde{x}_t|x_{<t})-\mathbb{E}_{\tilde{x}_t\sim p_t}\log P_{\theta}(\tilde{x}_t|x_{<t})=o(T).$$
\end{assumption}
\begin{assumption}
    \label{assumption 2}
    (Prescriptiveness of AI-generated text). There exists $\epsilon>0$, such that for any $t$,
    \begin{align*}
        &\mathbb{E}_{\tilde{x}_t\sim q_t}\log P_{\phi}(\tilde{x}_t|x_{<t})-\mathbb{E}_{\tilde{x}_t\sim p_t}\log P_{\phi}(\tilde{x}_t|x_{<t})\geq\epsilon,\\
        &\text{Var}_{\tilde{x}_t\sim q_t}\log P_{\phi}(\tilde{x}_t|x_{<t})\geq\text{Var}_{\tilde{x}_t\sim q_t}\Delta(\tilde{x}_t|x_{<t}).
    \end{align*}
\end{assumption}
\begin{assumption}
    \label{assumption 3}
    (Diversity of human-written text) 
    \begin{align*}
         \text{Cov}_{\tilde{x}_t\sim p_t}\left(\Delta(\tilde{x}_t|x_{<t}),\log P_{\phi}(\tilde{x}_t|x_{<t})\right)\geq  \text{Cov}_{\tilde{x}_t\sim q_t}\left(\Delta(\tilde{x}_t|x_{<t}),\log P_{\phi}(\tilde{x}_t|x_{<t})\right).
    \end{align*}
\end{assumption}

\Cref{assumption 1} formalizes the observation that the pre-trained base model is optimized to model the broad distribution of human language, making it statistically indistinguishable for both sources. 
\Cref{assumption 2} characterizes the phenomenon that the aligned model $P_\phi$ tends to yield higher likelihoods to AI-generated text.  \Cref{assumption 3}  suggests that due to the lexical and semantic divergence, under human-written text, the variance of the scores tends to be larger. Moreover, there is also a stronger positive correlation between the alignment shift and the aligned model's confidence. We further provide an empirical validation of these assumptions in Appendix~\ref{app:assumption-validation}.

Based on these assumptions, we first propose theoretical results of $S_{\text{Fast}}$ and the alignment imprint $\Delta(\boldsymbol{x})$. To compare at a fixed FNR, we consider the standardized $\Delta(\boldsymbol{x})$, which means $ S_{\text{Align}}=(\Delta(\boldsymbol{x}) - \tilde{\mu}_x)/\tilde{\sigma}_x$:
\begin{theorem}
    \label{theorem 1}
    Under \Cref{assumption 1} and \Cref{assumption 2}, for any fixed threshold $\tau$, we have $$\text{TNR}_{\tau}(S_{Align})\succeq \text{TNR}_{\tau}(S_{Fast}),$$
    where we use the notation $A\succeq B$ to denote that $A\geq B+o_{p}(1)$ as $T \rightarrow \infty$. In other words, the inequality holds asymptotically or with a negligible error term.
\end{theorem} 
\Cref{theorem 1} provides the theoretical justification for the alignment imprint, which effectively cancel out the intrinsic content-dependent component shared by the base model and the aligned model.

Furthermore, \Cref{theorem 2} below shows that under circumstances where $S_{Align}$ is not distinguishable, LAPD can significantly amplify the alignment signal and more effective.
\begin{theorem}
    \label{theorem 2}
    Denote $m_{\phi}=\sup_{\tilde{x}_t \sim q_t}|\log P_{\phi}(\tilde{x}_t|x_{<t})|,$ $ m_{\Delta}=\sup_{\tilde{x}_t \sim q_t}|\Delta(\tilde{x}_t|x_{<t})|.$ under \Cref{assumption 2} and \Cref{assumption 3}, if for some $k>0$,  $ k\cdot Var_{\tilde{x}_t \sim q_t}\Delta(\tilde{x}_t|x_{<t})\geq Var_{\tilde{x}_t \sim q_t}\log P_{\phi}(\tilde{x}_t|x_{<t})$, and   
    \begin{align*}
        \left|\frac{\mathbb{E}_{\tilde{x}_t\sim q_t}\Delta(\tilde{x}_t|x_{<t})}{\mathbb{E}_{\tilde{x}_t\sim p_t}\Delta(\tilde{x}_t|x_{<t})}-1\right|\leq \frac{\epsilon}{2m_{\phi}+\sqrt{k}m_{\Delta}},
    \end{align*}
     then we have $$\text{TNR}_{\tau}(S_{LAPD})\succeq \text{TNR}_{\tau}(S_{Align}).$$
    
\end{theorem}

The proofs of \Cref{theorem 1} and \Cref{theorem 2} can be seen in \Cref{appdendix: proof1} and \Cref{appdendix: proof2}, respectively.

\section{Experiments}
\subsection{Settings}

\textbf{Datasets.}
To evaluate the detectors in practical scenarios, we conduct our primary evaluations on four benchmarks: M4 \citep{wang2024m4}, DetectRL \citep{wu2024detectrl}, RAID \citep{dugan2024raid} and RealDet \citep{zhu2025realdet}, randomly select 2,000 balanced samples from each.
Additionally, we conduct on other three datasets: XSum \citep{narayan2018xsum}, WritingPrompts \citep{fan2018wp}, and Arxiv \citep{paul2021arxiv}. For each dataset, we collect 1,600 human-written and AI-generated text pairs, which are generated by three advanced LLMs: GPT-4-Turbo, Gemini-2.0-Flash, and Claude-3.7-Sonnet. 

\textbf{Baselines.}
We compare our \textit{Raw Alignment Imprint (RAI)} and \textit{Log-likelihood Alignment Preference Discrepancy (LAPD)} with 8 representative training-free detection methods, including Entropy \citep{ippolito2020entropy}, Log-Likelihood \citep{solaiman2019release}, LogRank \citep{gehrmann2019gltr}, DetectGPT \citep{mitchell2023detectgpt}, Fast-DetectGPT \citep{bao2024fast}, Lastde++ \citep{xu2025lastde}, Binoculars \citep{hans2024binoculars}, and DNA-DetectLLM \citep{zhu2025dna}.

\textbf{Evaluation Metrics.}
We adopt the commonly used area under the receiver operating characteristic (AUROC \citep{jimenez2012auroc}) curve as the main evaluation metric. 
Since the most harms often arise from false positives, we also adopt True Positive Rate (TPR) controlling False Positive Rate (FPR) at 0.5\%.

\textbf{Implementation.}
For RAI and LAPD, we use Llama2-7B \citep{touvron2023llama2} as base and perturbation model, and Llama2-7B-Instruct \citep{touvron2023llama2} as aligned model. More detailed information about implementation, benchmarks, datasets, and used LLMs can be seen in \Cref{appendix:exp}.


\subsection{Main Results}
\begin{table*}[h]
\caption{Detection performance (AUROC \%) on four comprehensive benchmarks. DetectRL Multi-LLM and DetectRL Multi-Domain are abbreviated as \textbf{Detect LLM} and \textbf{Detect Domain}, respectively. The best results are highlighted in \textbf{bold}. Methods with clubs ($\clubsuit$) standardize the scores by perturbing or generating auxiliary sequences.}
\vspace{-0.5em}
\label{tab:benchmark_results}
\begin{center}
\begin{small} 
\renewcommand{\arraystretch}{1.2} 
\begin{tabularx}{\textwidth}{lYYYYYY} 
\toprule
\textbf{Methods} & \textbf{M4} & \textbf{\makecell[c]{DetectRL \\ LLM}} & \textbf{\makecell[c]{DetectRL \\ Domain}} & \textbf{RAID} & \textbf{RealDet} & \textbf{Avg.} \\
\midrule
Entropy & 71.24 & 64.37 & 47.91 & 68.56 & 90.08 & 68.43 \\
Likelihood & 77.63 & 66.82 & 48.95 & 71.02 & 89.17 & 70.72 \\
LogRank & 78.69 & 67.29 & 50.52 & 72.03 & 89.93 & 71.69 \\
DetectGPT\perturb & 46.54 & 43.13 & 28.00 & 49.43 & 61.33 & 45.69 \\
Fast-DetectGPT\perturb & 81.10 & 82.13 & 74.81 & 80.59 & 92.66 & 82.26 \\
Lastde++\perturb & 79.44 & 75.18 & 66.87 & 82.03 & 92.39 & 79.18 \\
Binoculars & 83.42 & 83.26 & 77.50 & 82.30 & 93.67 & 84.03 \\
DNA-DetectLLM\perturb & 77.38 & 89.01 & 88.29 & 81.42 & 94.48 & 86.12 \\
\midrule
RAI & 70.46 & 93.13 & \textbf{96.19} & 81.28 & 86.92 & 85.60 \\
LAPD\perturb & \textbf{88.02} & \textbf{97.17} & 96.11 & \textbf{85.21} & \textbf{95.32} & \textbf{92.37} \\
\bottomrule
\end{tabularx}
\end{small}
\end{center}
\vspace{-0.5em}
\end{table*}
\begin{table*}[h]
\caption{Robustness across source LLMs and domains (AUROC \%). Source LLMs are abbreviated as: \textbf{GPT} (GPT-4 Turbo), \textbf{Gemini} (Gemini-2.0 Flash), and \textbf{Claude} (Claude-3.7 Sonnet). The best results are highlighted in \textbf{bold}. Methods with clubs ($\clubsuit$) standardize the scores by perturbing or generating auxiliary sequences.}
\vspace{-0.5em}
\label{tab:main_results}
\begin{center}
\begin{small}
\renewcommand{\arraystretch}{1.2}
\setlength{\tabcolsep}{0pt} 
\begin{tabularx}{\textwidth}{lYYYYYYYYYY} 
\toprule
\multirow{2}{*}{\textbf{Methods}} & \multicolumn{3}{c}{\textbf{XSum}} & \multicolumn{3}{c}{\textbf{WritingPrompts}} & \multicolumn{3}{c}{\textbf{Arxiv}} & \multirow{2}{*}{\textbf{Avg.}} \\
\cmidrule(lr){2-4} \cmidrule(lr){5-7} \cmidrule(lr){8-10}
 & GPT & Gemini & Claude & GPT & Gemini & Claude & GPT & Gemini & Claude & \\
\midrule
Entropy & 72.76 & 54.93 & 74.97 & 87.88 & 90.39 & 91.21 & 45.96 & 80.05 & 80.49 & 75.41 \\
Likelihood & 70.02 & 69.48 & 70.37 & 80.85 & 95.23 & 85.84 & 57.91 & 93.60 & 86.22 & 78.84 \\
LogRank & 69.58 & 69.25 & 69.76 & 78.92 & 94.53 & 84.12 & 58.19 & 94.15 & 85.80 & 78.26 \\
DetectGPT\perturb & 64.77 & 44.71 & 44.62 & 59.75 & 60.60 & 47.32 & 28.94 & 53.92 & 60.67 & 51.70 \\
Fast-DetectGPT\perturb & 98.17 & 91.41 & 93.32 & 95.89 & 98.01 & 91.83 & 91.57 & 99.80 & 98.21 & 95.36 \\
Lastde++\perturb & 96.59 & 88.13 & 89.98 & 91.85 & 96.45 & 84.41 & 88.97 & 99.54 & 97.47 & 92.60 \\
Binoculars & 98.13 & 95.68 & 96.94 & 97.79 & 99.55 & 97.36 & 93.74 & 99.87 & 98.05 & 97.46 \\
DNA-DetectLLM\perturb & 99.35 & 96.77 & 98.51 & 98.89 & 99.73 & 98.55 & 95.08 & 99.88 & 98.42 & 98.36 \\
\midrule
RAI & 98.94 & 94.77 & 95.94 & 94.50 & 96.25 & 87.85 & 95.94 & 99.18 & 97.85 & 95.69 \\
LAPD\perturb & \textbf{99.84} & \textbf{98.64} & \textbf{99.76} & \textbf{99.73} & \textbf{99.77} & \textbf{99.83} & \textbf{97.87} & \textbf{99.99} & \textbf{99.99} & \textbf{99.49} \\
\bottomrule
\end{tabularx}
\end{small}
\end{center}
\end{table*}

\Cref{tab:benchmark_results} shows the performance across four comprehensive benchmarks. 
RAI achieves a competitive average AUROC of $ 85.60\% $. LAPD further yields $ 92.37\% $, outperforming Fast-DetectGPT and the best baselines relatively by $56.99\%$ and $ 45.82\% $, where the ``relative'' is calculated by $(new-old)/(1.0-old)$, representing the improvement made relative to the maximum possible.

Notably, in the DetectRL Multi-Domain benchmark, RAI achieves the highest AUROC of 96.19\%. This observation indicates that our alignment imprint can effectively avoid interference from intrinsic content complexity, empirically validating the failure illustrated in \Cref{LAPD_framework}.A.

\subsection{Robustness Analysis}
\subsubsection{Robustness across LLMs and domains}
\Cref{tab:main_results} presents the detection results across various LLMs and domains. 
RAI achieves competitive performance without extra perturbations, while LAPD further improves the average AUROC by $ 89.01\% $ relative to Fast-DetectGPT. Even compared to the strongest baseline, LAPD achieves $ 68.90\% $ relative enhancement.
We further examine the transferability of a single detection threshold across domains in Appendix~\ref{app:cross-domain-threshold-transfer}.

\subsubsection{Robustness on Low-FPR Area}
In real-world scenarios, minimizing the rate of Type-I errors ($i.e.$ False Positive Rate, FPR) is often prioritized over aggregate metrics like AUROC. 
\Cref{fig:roc_curve} visually highlights this ability of LAPD, which maintains a high TPR (True Positive Rate) even when the FPR approaches zero, much better than all the baselines. 
Notably, the longer initial vertical segment indicates that TPR is highly sensitive to marginal changes in FPR, indicating a lack of robustness. While the baselines present significant vertical jumps, LAPD demonstrates superior stability in this low-FPR area.

To quantify this robustness across diverse domains, \Cref{tab:tpr_results} shows the TPR at a $ 0.5\% $ FPR threshold. 
Under this constraint, existing methods such as Fast-DetectGPT and DNA-DetectLLM exhibit significant performance drop. 
In contrast, our simple RAI already surpasses these baselines, demonstrating the intrinsic separability of the alignment signal. 
Furthermore, LAPD achieves an average TPR of $ 92.27\% $, representing a relative improvement of $76.81\%$, which highlights the robustness of LAPD under minimal false-positive tolerance. 


\subsubsection{Robustness on Different Text Lengths}
Detecting shorter texts is significantly challenging due to limited statistical signals \citep{verma2024ghostbuster}. 
To evaluate the robustness across varying text lengths, we truncate the input texts to target word counts. Specifically, we conduct experiments on the XSum dataset with AI text generated by GPT-4-Turbo, varying the text length from 20 to 200 words.

\begin{figure}[t]
    \centering
    \begin{minipage}[t]{0.48\linewidth}
        \centering
        \includegraphics[width=\linewidth]{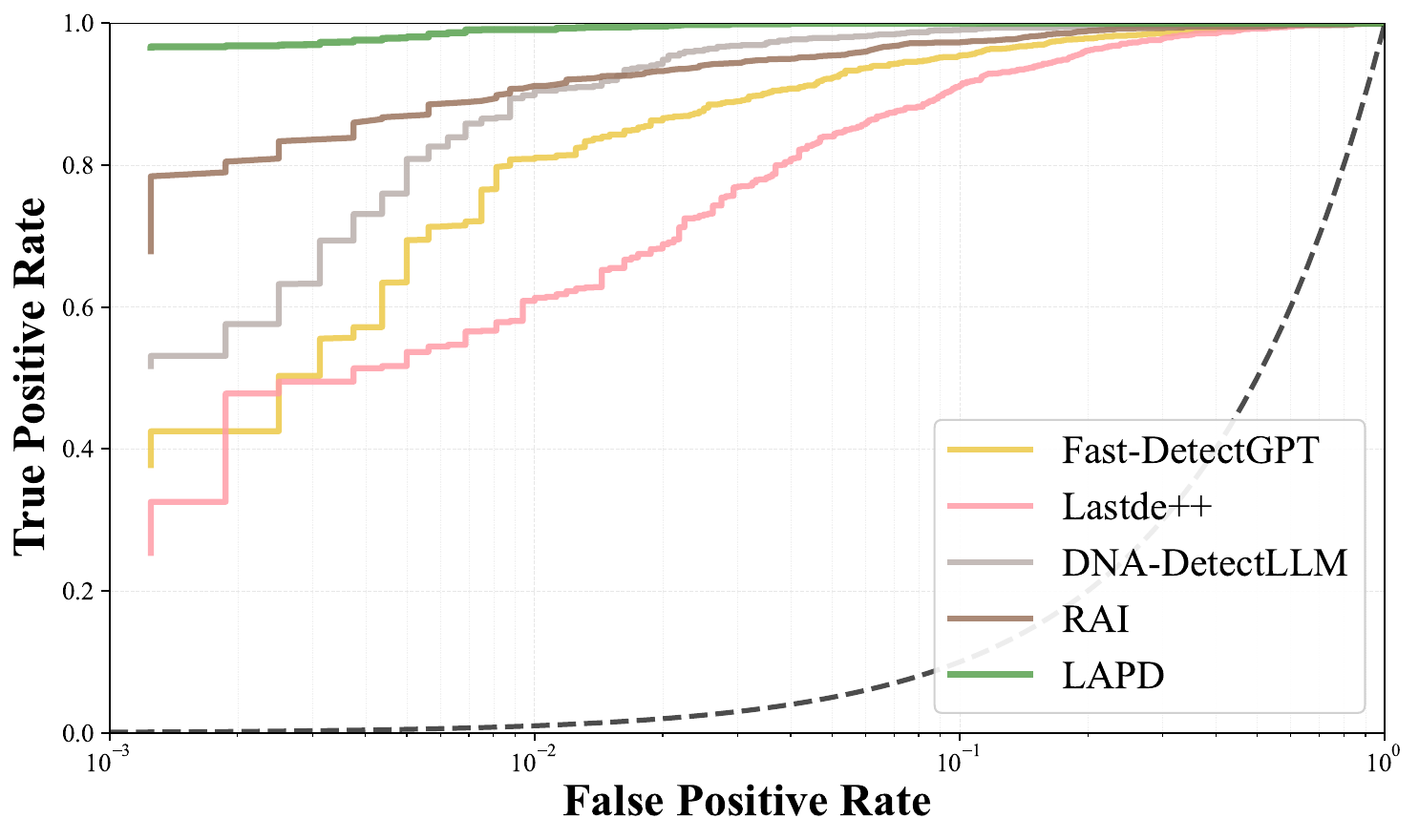}
        \vspace{-1.5em}
        \caption{ROC curve in log scale. The methods are evaluated on XSum dataset with AI text generated by GPT-4-Turbo. The dash line denotes the random classifier.}
        \label{fig:roc_curve}
    \end{minipage}
    \hfill
    \begin{minipage}[t]{0.48\linewidth}
        \centering
        \includegraphics[width=\linewidth]{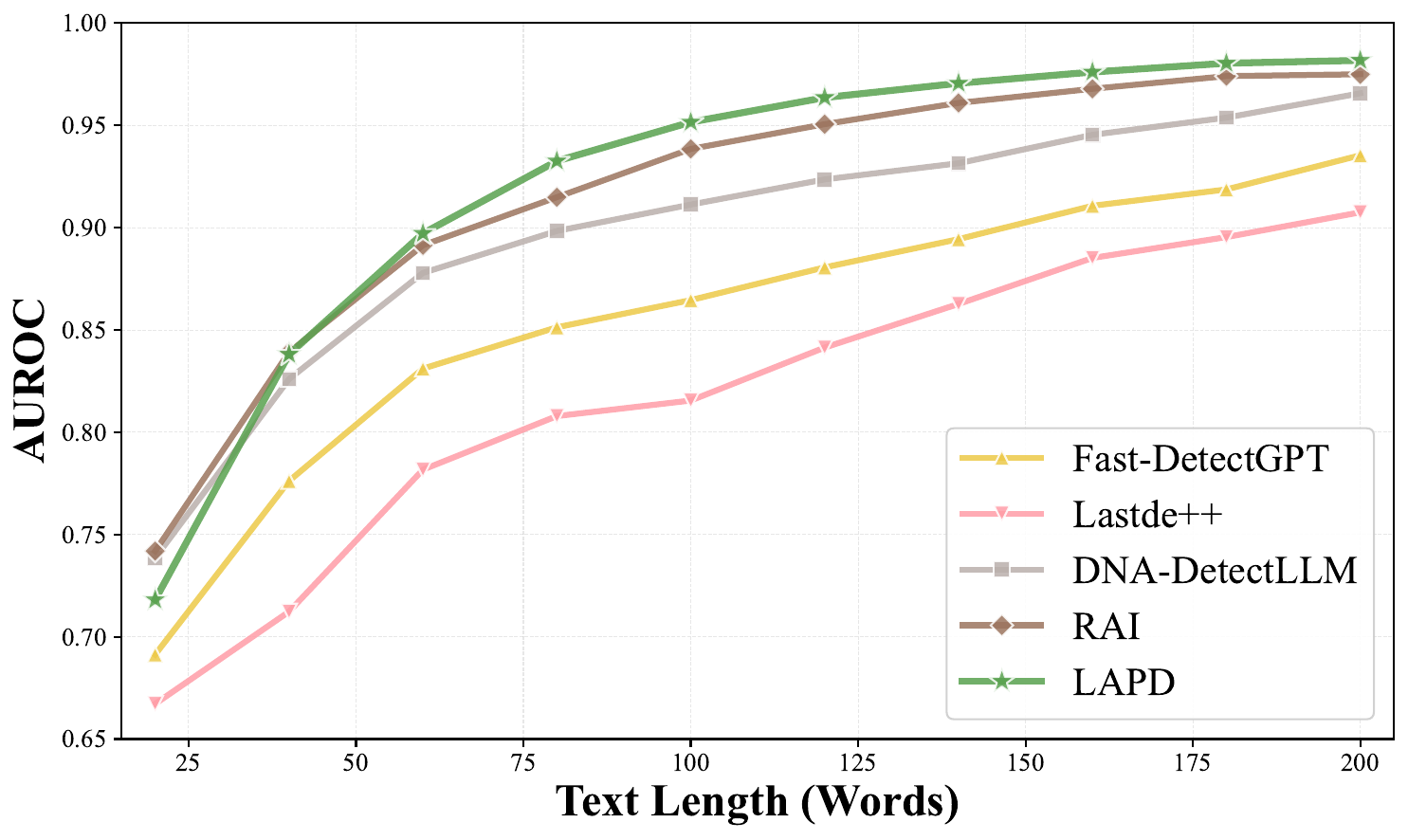}
        \vspace{-1.5em}
        \caption{Detection performance (AUROC \%) across varying text lengths (20--200 words). The methods are evaluated on XSum dataset with AI text generated by GPT-4-Turbo.}
        \label{fig:text_length_ablation}
    \end{minipage}
    \vspace{-1.5em}
\end{figure}

\Cref{fig:text_length_ablation} visualizes the performance curves of five representative methods. While all detectors show improved AUROC as the length increases, LAPD demonstrates the most rapid AUROC growth in the short-text interval (20-80 words).
Moreover, LAPD consistently outperforms all baselines once the length exceeds 40 words, while RAI ranking second. This observation presents the intrinsic robustness of our methods across various text lengths.

\subsubsection{Robustness under Attacks}
We apply three types of token-level attacks: insertion, deletion, and replacement to 1\% of the tokens and further extend to four more realistic attacks: paraphrase, synonym substitution, homoglyph, and zero-width space insertion from the RAID benchmark. Overall, our LAPD method outperforms all baselines, achieving the highest average attack AUROC and the highest clean performance, as illustrated in  \Cref{appendix:realistic_attacks}.



\subsection{Ablation Studies}
\subsubsection{Component Analysis}
We evaluate the individual contribution of each component of LAPD, as summarized in \Cref{tab:ablation_components}. 
The results of log-likelihood alone are poor, regardless of base model or aligned model. This can be attributed to their high sensitivity to intrinsic content complexity.
$\Delta$ achieves superior results by capturing the alignment imprint. 
Compared with $\Delta$, $\mathcal{S}$ exhibits comparable performance on XSum and attains relative gains of $ 41.65\% $ and $ 16.24\% $ on WritingPrompts and Arxiv, respectively. These improvements confirm the suboptimality of $\Delta $ and the utility of the information-weighted score.
Finally, LAPD outperforms all baselines across the three datasets, demonstrating its effectiveness.

\subsubsection{Other Base-Aligned Model Pairs}
\Cref{tab:ablation_model_pairs_small} evaluates the sensitivity of LAPD to various model pairs.
While the Llama2-7B pair has the best performance, the Falcon-7B, GPT-J-6B, and Llama3.1-8B pairs also demonstrate robust performance, achieving average AUROCs of $ 89.01\% $, $ 92.98\% $, and $ 96.71\% $, respectively.

\begin{table}[t]
\begin{minipage}[t]{0.48\linewidth}
    \centering
    \captionof{table}{Ablation results (AUROC\,\%) of different
    components of LAPD. 
    Best results in \textbf{bold}, second \underline{underlined}.}
    \label{tab:ablation_components}
    \vspace{0.4em}
    \small
    \begin{tabular}{lccc}
        \toprule
        \textbf{Statistic} & \textbf{XSum} & \textbf{Writing} & \textbf{Arxiv} \\
        \midrule
        $\log P_\theta$          & 69.57 & 89.23 & 82.07 \\
        $\log P_\phi$            & 69.96 & 87.31 & 79.24 \\
        $\Delta$                 & \underline{96.55} & 92.87 & 97.66 \\
        $\mathcal{S}$            & 96.40 & \underline{95.84} & \underline{98.04} \\
        \midrule
        $\mathcal{S}_\text{LAPD}$ & \textbf{99.42} & \textbf{99.78} & \textbf{99.28} \\
        \bottomrule
    \end{tabular}
\end{minipage}
\hfill
\begin{minipage}[t]{0.48\linewidth}
    \centering
    \captionof{table}{Ablation results (AUROC\,\%) of LAPD with
    different model pairs across three datasets.
    Llama2-7B pair serves as the default.}
    \label{tab:ablation_model_pairs_small}
    \vspace{0.4em}
    \small
    \begin{tabular}{lccc}
        \toprule
        \textbf{Model Pairs} & \textbf{XSum} & \textbf{Writing} & \textbf{Arxiv} \\
        \midrule
        Falcon-7B Pair   & 87.35 & 93.37 & 86.29 \\
        GPT-J-6B Pair    & 94.62 & 91.56 & 92.75 \\
        Llama3.1-8B Pair & 97.09 & 94.26 & 98.78 \\
        \midrule
        Llama2-7B Pair   & \textbf{99.42} & \textbf{99.54} & \textbf{99.61} \\
        \bottomrule
    \end{tabular}
\end{minipage}
\end{table}

\subsection{Time Efficiency}
Beyond detection accuracy, computational efficiency is a critical factor for AI-generated text detection, especially in real-time scenarios.
\Cref{tab:time_efficiency} reports the time cost per sample for each method. To ensure a rigorous comparison, we detected 300 instances from the benchmarks and measured the average processing time with a batch size of 1.

While classic methods like Entropy, Likelihood, and LogRank are efficient ($\approx$ 0.3s), their practical utility is limited by bad performance.
The other methods process one sample range from 0.35s to 0.80s, among which RAI is the fastest and LAPD achieves the best performance.

\vspace{-1em}

\begin{table}[ht]
\caption{Per-text time costs (s) of all the methods. Methods with clubs ($\clubsuit$) standardize the scores by perturbing or generating auxiliary sequences.}
\label{tab:time_efficiency}
\begin{center}
\begin{small}
\renewcommand{\arraystretch}{1.2}
\setlength{\tabcolsep}{8pt} 
\begin{tabular}{lclc}
\toprule
\textbf{Methods} & \textbf{Time (s)} & \textbf{Methods} ($\clubsuit$) & \textbf{Time (s)} \\
\midrule
Entropy     & 0.2784 & DetectGPT\perturb      & 17.8047 \\
Likelihood  & 0.2777 & Fast-DetectGPT\perturb & 0.5578 \\
LogRank     & 0.2851 & Lastde++\perturb       & 0.6169 \\
Binoculars  & 0.6549 & DNA-DetectLLM\perturb  & 0.7696 \\
\midrule
RAI         & 0.3506 & LAPD\perturb           & 0.5792 \\
\bottomrule
\end{tabular}
\end{small}
\end{center}
\vspace{-1.5em}
\end{table}

\section{Conclusion}

In this paper, we introduce a novel perspective for AI-generated text detection by characterizing the distribution shift induced by LLM alignment. We theoretically formalize it as the \textit{Alignment Imprint}, demonstrating that the log-likelihood ratio can naturally disentangle alignment-induced biases from intrinsic content complexity. Furthermore, we propose the \textit{\textbf{L}og-likelihood \textbf{A}lignment \textbf{P}reference \textbf{D}iscrepancy (LAPD)}, a standardized information-weighted statistic.
We provide statistical guarantee that LAPD dominates Fast-DetectGPT. We also theoretically show that LAPD improves further when the aligned and base models are close in distribution. Extensive experiments show that LAPD achieves an improvement $45.82\%$ relative to the strongest existing baselines and exhibits exceptional robustness in the low-FPR area.

\subsection*{ACKNOWLEDGEMENT}
This work is supported in part by the National Natural Science Foundation of China under grant 62301189, 62576122,62571298, Guangdong Basic and Applied Basic Research Foundation under grant 2026A1515011139. We also thank all anonymous reviewers and the area chair for their helpful comments. 
\bibliographystyle{plainnat}
\bibliography{neurips_2026}

\crefalias{section}{appendix}
\crefalias{subsection}{appendix}
\appendix

\section{Derivations of the Alignment Distributions}

\subsection{Derivation in the SFT stage}
\label{appendix:sft_derivation}

In this section, we derive the theoretical form of the SFT distribution under the maximum entropy framework. Without loss of generality, the goal is to find a data distribution $P(\boldsymbol{x})$ that maximizes the expected latent quality $V(\boldsymbol{x})$ (representing adherence to instructions and format) while remaining close to the pre-trained base distribution $P_\theta(\boldsymbol{x})$ to preserve linguistic capabilities and knowledge. Formally, we frame this as a constrained optimization problem:
\begin{equation}
\min_P D_{\text{KL}}(P \parallel P_\theta) \quad \text{s.t.} \quad \mathbb{E}_{\boldsymbol{x} \sim P}[V(\boldsymbol{x})] \geq \mu,
\end{equation}

where $\mu$ is a target quality threshold. The Lagrangian for this optimization problem is:
\begin{equation}
\mathcal{L}(P, \lambda, \gamma) =\sum_{\boldsymbol{x}} P(\boldsymbol{x}) \log \frac{P(\boldsymbol{x})}{P_\theta(\boldsymbol{x})} + \gamma \left( \sum_{\boldsymbol{x}} P(\boldsymbol{x}) - 1 \right) - \lambda \left( \sum_{\boldsymbol{x}} P(\boldsymbol{x}) V(\boldsymbol{x}) - \mu \right),
\end{equation}

where $\lambda$ is the Lagrange multiplier associated with the quality constraint (related to the inverse temperature), and $\gamma$ enforces the normalization constraint $\sum P(\boldsymbol{x}) = 1$. 

To find the optimal distribution $P^*$, we take the functional derivative with respect to $P(\boldsymbol{x})$ and set it to zero:
\begin{equation}
\log P^*(\boldsymbol{x}) + 1 - \log P_\theta(\boldsymbol{x}) - \lambda V(\boldsymbol{x}) + \gamma = 0.
\end{equation}

Rearranging the equation and exponentiating both sides, we can obtain as follow:
\begin{equation}
P^*(\boldsymbol{x}) = P_\theta(\boldsymbol{x}) \exp(\lambda V(\boldsymbol{x})) \exp(-(1+\gamma)).
\end{equation}

To satisfy the normalization constraint $\sum_{\boldsymbol{x}} P(\boldsymbol{x}) = 1$, the term $\exp(-(1+\gamma))$ must equal the inverse of the partition function. Let us define the partition function $Z_1$ as:
\begin{equation}
Z_1 = \sum_{\boldsymbol{x}'} P_\theta({\boldsymbol{x}'}) \exp(\lambda V({\boldsymbol{x}'})).
\end{equation}

Absorbing the constants $\lambda$ (by redefining the scale of $V$) and $\exp(-(1+\gamma))$ into a normalization factor $1/Z_1$, we arrive at the Boltzmann distribution:
\begin{equation}
P^*(\boldsymbol{x}) = \frac{1}{Z_1} P_\theta(\boldsymbol{x}) \exp(V(\boldsymbol{x})).
\end{equation}

The training objective of SFT is maximum likelihood estimation. Although practical models $P_\text{SFT}$ may not reach optimality, the probability mass of $P_{\text{SFT}}(\boldsymbol{x})$ will be concentrated in regions where $P_{\text{SFT}}^*(\boldsymbol{x})$ is high. Therefore, $P_{\text{SFT}}$ is monotonically correlated with the latent quality function $V(\boldsymbol{x})$.

\subsection{Derivation in the preference tuning stage}
\label{appendix:rl_derivation}

Here we derive the closed-form solution for the preference tuning stage. The objective is to maximize the expected reward $R(\boldsymbol{x})$ subject to a KL-divergence constraint relative to the reference policy (which is usually $P_{\text{SFT}}$ in the LLM alignment pipeline):
\begin{equation}
\max_\phi \mathcal{J}(\phi) = \mathbb{E}_{\boldsymbol{x} \sim P_\phi}[R(\boldsymbol{x})] - \beta D_{\text{KL}}(P_\phi \parallel P_{\text{ref}}).
\end{equation}

Expanding the KL divergence term:
\begin{equation}
\begin{split}
\mathcal{J}(\phi) &= \sum_{\boldsymbol{x}} P_\phi(\boldsymbol{x}) R(\boldsymbol{x}) - \beta \sum_{\boldsymbol{x}} P_\phi(\boldsymbol{x}) \log \frac{P_\phi(\boldsymbol{x})}{P_{\text{ref}}(\boldsymbol{x})} \\
&= \sum_{\boldsymbol{x}} P_\phi(\boldsymbol{x}) \left( R(\boldsymbol{x}) - \beta \log P_\phi(\boldsymbol{x}) + \beta \log P_{\text{ref}}(\boldsymbol{x}) \right).
\end{split}
\end{equation}

To find the optimal policy $P_{\phi^*}$, we introduce a Lagrange multiplier $\gamma$ for the normalization constraint $\sum P_\phi(\boldsymbol{x}) = 1$:
\begin{equation}
\begin{split}
\mathcal{L}(P_\phi, \gamma) = \mathcal{J}(\phi) - \gamma \left( \sum_{\boldsymbol{x}} P_\phi(\boldsymbol{x}) - 1 \right).
\end{split}
\end{equation}

Taking the derivative with respect to $P_\phi(\boldsymbol{x})$ and setting to zero:
\begin{align}
R(\boldsymbol{x}) - \beta (\log P_{\phi^*}(\boldsymbol{x}) + 1) + \beta \log P_{\text{ref}}(\boldsymbol{x}) - \gamma = 0.
\end{align}

Rearranging the equation and dividing by $\beta$:
\begin{equation}
\log P_{\phi^*}(\boldsymbol{x}) - \log P_{\text{ref}}(\boldsymbol{x}) = \frac{1}{\beta} R(\boldsymbol{x}) - \log Z_2,
\end{equation}

where $\log Z_2=\frac{\gamma + \beta}{\beta}$ is a constant with respect to $\boldsymbol{x}$. Let us solve for it using the constraint $\sum P_{\phi^*} = 1$. From this equation, we can write the optimal policy $P_{\phi^*}(\boldsymbol{x})$ as:
\begin{equation}
P_{\phi^*}(\boldsymbol{x}) = \frac{1}{Z_2}P_{\text{ref}}(\boldsymbol{x}) \exp\left( \frac{R(\boldsymbol{x})}{\beta} \right).
\end{equation}

Summing over all $\boldsymbol{x}'$:
\begin{equation}
1 = \sum_{\boldsymbol{x}'} P_\phi^*(\boldsymbol{x}') = \frac{1}{Z_2} \sum_{\boldsymbol{x}'} P_{\text{ref}}(\boldsymbol{x}') \exp\left( \frac{R(\boldsymbol{x}')}{\beta} \right).
\end{equation}

Then we obtain the constant 
\begin{equation}
Z_2=\sum_{\boldsymbol{x}'} P_{\text{ref}}(\boldsymbol{x}') \exp\left( \frac{R(\boldsymbol{x}')}{\beta} \right).
\end{equation}

\section{Proofs}
To begin, we first present two technical assumptions, which is analogous to the setting introduced in \citep{zhou2025adadetectgpt}. 
\begin{assumption}
\label{assumption 4}
    Define \begin{align*}
    \sigma_{T,S_{\text{LAPD}}}=\frac{\sqrt{\sum_{t=1}^T\text{Var}_{\tilde{x}_t\sim q_t}S (\tilde{x}_t|x_{<t})}}{\sqrt{\sum_{t=1}^T\text{Var}_{\tilde{x}_t\sim p_t}S (\tilde{x}_t|x_{<t})}}.
    \end{align*}
    Similarly, we define $\sigma_{T,S_{\text{Align}}}$ and $\sigma_{T,S_{\text{Align}}}$. Then $\sigma_{T,S_{\text{LAPD}}}-\sigma_{T,S_{\text{Align}}}\rightarrow0, \ \sigma_{T,S_{\text{Align}}}-\sigma_{T,S_{\text{Align}}}\rightarrow0,$ in probability.
\end{assumption}
\begin{assumption}
\label{assumption 5}
    Define \begin{align*}
    \bar{\sigma}_{T,S_{\text{LAPD}}}=\frac{\sqrt{\sum_{t=1}^T\text{Var}_{\tilde{x}_t\sim q_t}S (\tilde{x}_t|x_{<t})}}{\sqrt{\sum_{t=1}^T\text{Var}_{\boldsymbol{\tilde{x}}\sim \boldsymbol{q}}S (\tilde{x}_t|\tilde{x}_{<t})}}.
    \end{align*}
    Similarly, we define $\bar{\sigma}_{T,S_{\text{Align}}}$ and $\bar{\sigma}_{T,S_{\text{Align}}}$.  If $\boldsymbol{x}\sim \boldsymbol{q}$, then $\bar{\sigma}_{T,S_{\text{LAPD}}}\rightarrow 1,\ \bar{\sigma}_{T,S_{\text{Align}}}\rightarrow 1, \ \bar{\sigma}_{T,S_{\text{Align}}} \rightarrow 1,$ in probability.
\end{assumption}
It is worth noting that, Assumption 3 in \citep{zhou2025adadetectgpt} basically
requires the conditional variance of logits be asymptotically equivalent for human-written text and AI-generated text. However, this may not hold in real-world scenarios since human-written text tends to exhibit higher variance due to lexical and semantic diversity. In contrast, our \Cref{assumption 4} only requires the ratio of the sentence-level variance under AI-generated  and human-written text to be asymptotically equivalent across the three statistics.

The following \Cref{lemma 1} parallels the Lemma S1 in \citep{zhou2025adadetectgpt}, which can be viewed as a non-asymptotic version of martingale central limit theorem. \Cref{lemma 1} is essential for proving our main theorems.
\begin{lemma}
\label{lemma 1}
    Let $\boldsymbol{x} = (x_1, \dots x_n)$ be sequences of real valued random variables satisfying for all $1 \le t \le n$,
\[
\mathbb{E}(x_t | x_{<t}) = 0 \quad \text{almost surely.}
\]
Let $\sigma_t^2 = \mathbb{E}(x_t^2 | x_{<t})$, $\bar{\sigma}_t^2 = \mathbb{E}(x_t^2)$, $s_n^2 = \sum_{t=1}^n \bar{\sigma}_t^2$ and $v_n^2 = \sum_{t=1}^n \sigma_t^2 / s_n^2$. Suppose $|x_n|$ is bounded by some constant almost surely for all $n$ and $s_n / \sqrt{n}$ is bounded away from zero. Then
\[
\sup_{z \in \mathbb{R}} \left| \mathbb{P} \left( \frac{\sum_{t=1}^n x_t}{\sqrt{\sum_{t=1}^n \sigma_t^2}} \le z \right) - \Phi(z) \right| = O \left( \frac{\log n}{\sqrt{n}} + (\mathbb{E}|v_n^2 - 1|)^{1/3} \right),
\]
where $\Phi(\bullet)$ is the cumulative distribution function of standard normal distribution.
\end{lemma}

\subsection{Proof of \Cref{theorem 1}}
\label{appdendix: proof1}
By \Cref{lemma 1}, we obtain that
\begin{align*}
    \text{TNR}_{\tau}(S_{\text{Align}})
    &=\mathbb{P}_{x\sim p}(S_{\text{Align}}(x)\leq \tau) \\
    &=\mathbb{P}_{x\sim p}\left(
        \frac{\sum_{t=1}^T \Delta (\tilde{x}_t|x_{<t})-\sum_{t=1}^T \mathbb{E}_{\tilde{x}_t\sim q_t}\Delta (\tilde{x}_t|x_{<t})}{\sqrt{\sum_{t=1}^T\text{Var}_{\tilde{x}_t\sim q_t}\Delta (\tilde{x}_t|x_{<t})}}
        \leq \tau
    \right)  \\
    &=\mathbb{P}_{x\sim p}\left(
        \frac{\sum_{t=1}^T \Delta (\tilde{x}_t|x_{<t})-\sum_{t=1}^T \mathbb{E}_{\tilde{x}_t\sim q_t}\Delta (\tilde{x}_t|x_{<t})}{\sqrt{\sum_{t=1}^T\text{Var}_{\tilde{x}_t\sim q_t}\Delta (\tilde{x}_t|x_{<t})}}
        \cdot \sigma_{T,S_{\text{Align}}}
        \leq \tau \cdot \sigma_{T,S_{\text{Align}}}
    \right)  \\
    &=\mathbb{P}_{x\sim p}\left(
        \frac{\sum_{t=1}^T \Delta (\tilde{x}_t|x_{<t})-\sum_{t=1}^T \mathbb{E}_{\tilde{x}_t\sim p_t}\Delta (\tilde{x}_t|x_{<t})}{\sqrt{\sum_{t=1}^T\text{Var}_{\tilde{x}_t\sim p_t}\Delta (\tilde{x}_t|x_{<t})}}
        \leq \tau\cdot \sigma_{T,S_{\text{Align}}} + \right. \\
    &\phantom{=\mathbb{P}_{x\sim p}\left( \right.}  
        \left. \frac{ \sum_{t=1}^T \mathbb{E}_{\tilde{x}_t\sim q_t}\Delta (\tilde{x}_t|x_{<t})-\sum_{t=1}^T \mathbb{E}_{\tilde{x}_t\sim p_t}\Delta (\tilde{x}_t|x_{<t})}{\sqrt{\sum_{t=1}^T\text{Var}_{\tilde{x}_t\sim q_t}\Delta (\tilde{x}_t|x_{<t})}}
    \right) \\
    &= \mathbb{E}_{x\sim p} \Phi \left(
        \tau\cdot \sigma_{T,S_{\text{Align}}} + \frac{ \sum_{t=1}^T \mathbb{E}_{\tilde{x}_t\sim q_t}\Delta (\tilde{x}_t|x_{<t})-\sum_{t=1}^T \mathbb{E}_{\tilde{x}_t\sim p_t}\Delta (\tilde{x}_t|x_{<t})}{\sqrt{\sum_{t=1}^T\text{Var}_{\tilde{x}_t\sim q_t}\Delta (\tilde{x}_t|x_{<t})}}
    \right) + o_{p}(1)  \tag{23}
\end{align*}
The last equality is obtained by \Cref{lemma 1} and \Cref{assumption 5}. Similarly, we have
\begin{align*}
    \text{TNR}_{\tau}(S_{Align})=\mathbb{E}_{x\sim p} \Phi \left(\tau\cdot \sigma_{T,S_{\text{Align}}}+\frac{ \sum_{t=1}^T \mathbb{E}_{\tilde{x}_t\sim q_t}\log P_{\phi} (\tilde{x}_t|x_{<t})-\sum_{t=1}^T \mathbb{E}_{\tilde{x}_t\sim p_t}\log P_{\phi} (\tilde{x}_t|x_{<t})}{\sqrt{\sum_{t=1}^T\text{Var}_{\tilde{x}_t\sim q_t}  \log P_{\phi}(\tilde{x}_t|x_{<t})}}\right)+o_{p}(1).
\end{align*}
By \Cref{assumption 1,assumption 2}, we obtain that:
\begin{align*}
    &\mathbb{E}_{x\sim p} \Phi \left(\tau\cdot \sigma_{T,S_{\text{Align}}}+ \frac{ \sum_{t=1}^T \mathbb{E}_{\tilde{x}_t\sim q_t}\Delta (\tilde{x}_t|x_{<t})-\sum_{t=1}^T \mathbb{E}_{\tilde{x}_t\sim p_t}\Delta (\tilde{x}_t|x_{<t})}{\sqrt{\sum_{t=1}^T\text{Var}_{\tilde{x}_t\sim q_t}\Delta (\tilde{x}_t|x_{<t})}}\right)\\
    &\geq \mathbb{E}_{x\sim p} \Phi \left(\tau\cdot \sigma_{T,S_{\text{Align}}}+\frac{ \sum_{t=1}^T \mathbb{E}_{\tilde{x}_t\sim q_t}\Delta (\tilde{x}_t|x_{<t})-\sum_{t=1}^T \mathbb{E}_{\tilde{x}_t\sim p_t}\Delta (\tilde{x}_t|x_{<t})}{\sqrt{\sum_{t=1}^T\text{Var}_{\tilde{x}_t\sim q_t}  \log P_{\phi}(\tilde{x}_t|x_{<t})}}\right)\\
    &=\mathbb{E}_{x\sim p} \Phi \left(\tau\cdot \sigma_{T,S_{\text{Align}}}+\frac{ \sum_{t=1}^T \mathbb{E}_{\tilde{x}_t\sim q_t}\log P_{\phi} (\tilde{x}_t|x_{<t})-\sum_{t=1}^T \mathbb{E}_{\tilde{x}_t\sim q_t}\log P_{\phi} (\tilde{x}_t|x_{<t})}{\sqrt{\sum_{t=1}^T\text{Var}_{\tilde{x}_t\sim p_t} \log P_{\phi}(\tilde{x}_t|x_{<t})}}\right)+o_{p}(1), \tag{24}\\ 
\end{align*}
where the last equality arises from $\sigma_{T,S_{\text{Align}}}-\sigma_{T,\Delta} =o_{p}(1)$ in \Cref{assumption 4} and $\sum_{t=1}^T \mathbb{E}_{\tilde{x}_t\sim q_t}\log P_{\theta}(\tilde{x}_t|x_{<t})-\mathbb{E}_{\tilde{x}_t\sim p_t}\log P_{\theta}(\tilde{x}_t|x_{<t})=o(T)$ in \Cref{assumption 1}. This completes the proof.

\subsection{Proof of \Cref{theorem 2}}
\label{appdendix: proof2}
Following the same proof strategy as in \Cref{theorem 1}, we have 
\begin{align*}
    \text{TNR}_{\tau}(S_{\text{LAPD}})=\mathbb{E}_{x\sim p} \Phi \left(\tau\cdot \sigma_{T,S_{\text{LAPD}}}+\frac{ \sum_{t=1}^T \mathbb{E}_{\tilde{x}_t\sim p_t}S (\tilde{x}_t|x_{<t})-\sum_{t=1}^T \mathbb{E}_{\tilde{x}_t\sim q_t}S (\tilde{x}_t|x_{<t})}{\sqrt{\sum_{t=1}^T\text{Var}_{\tilde{x}_t\sim q_t}  S_{\phi}(\tilde{x}_t|x_{<t})}}\right)+o_{p}(1).
\end{align*}
 We first prove that \begin{align*}
    \text{Var}_{\tilde{x}_t\sim q_t}S_{\phi}(\tilde{x}_t|x_{<t})\leq (2m_{\phi}+\sqrt{k}m_{\Delta})^2 \cdot \text{Var}_{\tilde{x}_t\sim q_t}\Delta(\tilde{x}_t|x_{<t}).
\end{align*}
For simplicity, let $p_t=-\log P_{\phi}(\tilde{x}_t|x_{<t}), \Delta_t=\Delta(\tilde{x}_t|x_{<t})$.
Note that for random variables $X,Y,Z$, 
\begin{align*}
    \text{Var}(X+Y+Z)&= \text{Var}(X)+\text{Var}(Y)+\text{Var}(Z)+2\text{Cov}(X,Y)+2\text{Cov}(X,Z)+2\text{Cov}(Y,Z)\\
    &\leq \text{Var}(X)+\text{Var}(Y)+\text{Var}(Z)+2\sqrt{\text{Var}(X)\text{Var}(Y)}\\
    &\quad +2\sqrt{\text{Var}(X)\text{Var}(Z)}+2\sqrt{\text{Var}(Y)\text{Var}(Z)}\\
    &=(\sqrt{\text{Var}(X)}+\sqrt{\text{Var}(Y)}+\sqrt{\text{Var}(Z)})^2.
\end{align*}
Hence,
 \begin{align*}
    \sqrt{\text{Var}_{\tilde{x}_t\sim q_t}S_{\phi}(\tilde{x}_t|x_{<t})}&= \sqrt{\text{Var} \left[p_t(\Delta_t-\mathbb{E}\Delta_t)+\Delta_t(p_t-\mathbb{E}p_t)-(\Delta_t-\mathbb{E}\Delta_t)(p_t-\mathbb{E}p_t)\right]}\\
    &\leq \sqrt{\text{Var}\left[p_t(\Delta_t-\mathbb{E}\Delta_t)\right]}+\sqrt{\text{Var}\left[\Delta_t(p_t-\mathbb{E}p_t)\right]}\\
    &\quad +\sqrt{\text{Var}\left[(\Delta_t-\mathbb{E}\Delta_t)(p_t-\mathbb{E}p_t)\right]}\\
    &\leq  m_{\phi}\sqrt{\text{Var}\Delta_t}+m_{\Delta}\sqrt{\text{Var}\ p_t}+\sqrt{\mathbb{E}(\Delta_t-\mathbb{E}\Delta_t)^2(p_t-\mathbb{E}p_t)^2}\\
    &\leq (2m_{\phi}+\sqrt{k}m_{\Delta})\text{Var} \Delta_t=(2m_{\phi}+\sqrt{k}m_{\Delta})\sqrt{\text{Var}_{\tilde{x}_t\sim q_t}\Delta(\tilde{x}_t|x_{<t})}, \tag{25}
\end{align*}
where the last inequality follows from $0\leq p_t\leq m_{\phi},$ hence $|p_t-\mathbb{E}p_t|^2\leq m^2_{\phi}$, almost surely. 

\Cref{assumption 3} implies that
\begin{align*}
     &\mathbb{E}_{\tilde{x}_t\sim p_t}S (\tilde{x}_t|x_{<t})- \mathbb{E}_{\tilde{x}_t\sim p_t}\log P_{\phi}(\tilde{x}_t|x_{<t})\cdot\mathbb{E}_{\tilde{x}_t\sim p_t}\Delta(\tilde{x}_t|x_{<t})\\ &\geq \mathbb{E}_{\tilde{x}_t\sim q_t}S (\tilde{x}_t|x_{<t})  -\mathbb{E}_{\tilde{x}_t\sim q_t}\log P_{\phi}(\tilde{x}_t|x_{<t})\cdot\mathbb{E}_{\tilde{x}_t\sim q_t}\Delta(\tilde{x}_t|x_{<t}),
\end{align*}
then we have 
\begin{align*}
    &\frac{ \sum_{t=1}^T \mathbb{E}_{\tilde{x}_t\sim p_t}S (\tilde{x}_t|x_{<t})-\sum_{t=1}^T \mathbb{E}_{\tilde{x}_t\sim q_t}S (\tilde{x}_t|x_{<t})}{\sqrt{\sum_{t=1}^T\text{Var}_{\tilde{x}_t\sim q_t}  S_{\phi}(\tilde{x}_t|x_{<t})}}\\
    &\geq\frac{ \sum_{t=1}^T \mathbb{E}_{\tilde{x}_t\sim p_t}\log P_{\phi}(\tilde{x}_t|x_{<t})\cdot\mathbb{E}_{\tilde{x}_t\sim p_t}\Delta(\tilde{x}_t|x_{<t})-\sum_{t=1}^T\mathbb{E}_{\tilde{x}_t\sim q_t}\log P_{\phi}(\tilde{x}_t|x_{<t})\cdot\mathbb{E}_{\tilde{x}_t\sim q_t}\Delta(\tilde{x}_t|x_{<t})}{(2m_{\phi}+\sqrt{k}m_{\Delta})\sqrt{\sum_{t=1}^T\text{Var}_{\tilde{x}_t\sim q_t}\Delta (\tilde{x}_t|x_{<t})}}  \\
    &\geq \frac{\sum_{t=1}^T \mathbb{E}_{\tilde{x}_t\sim p_t}\Delta(\tilde{x}_t|x_{<t})\cdot \left(\mathbb{E}_{\tilde{x}_t\sim p_t}\log P_{\phi}(\tilde{x}_t|x_{<t})-\mathbb{E}_{\tilde{x}_t\sim q_t}\log P_{\phi}(\tilde{x}_t|x_{<t})\right)}{(2m_{\phi}+\sqrt{k}m_{\Delta})\sqrt{\sum_{t=1}^T\text{Var}_{\tilde{x}_t\sim q_t}\Delta (\tilde{x}_t|x_{<t})}} \\
    &\geq \frac{ \sum_{t=1}^T \mathbb{E}_{\tilde{x}_t\sim q_t}\Delta (\tilde{x}_t|x_{<t})-\sum_{t=1}^T \mathbb{E}_{\tilde{x}_t\sim p_t}\Delta (\tilde{x}_t|x_{<t})}{\sqrt{\sum_{t=1}^T\text{Var}_{\tilde{x}_t\sim q_t}\Delta (\tilde{x}_t|x_{<t})}},  \tag{26}
\end{align*}
where the last inequality follows from the assumption that
 \begin{align*}
        \frac{\mathbb{E}_{\tilde{x}_t\sim p_t}\Delta(\tilde{x}_t|x_{<t})-\mathbb{E}_{\tilde{x}_t\sim q_t}\Delta(\tilde{x}_t|x_{<t})}{\mathbb{E}_{\tilde{x}_t\sim p_t}\Delta(\tilde{x}_t|x_{<t})}&\leq  \frac{\epsilon}{2m_{\phi}+\sqrt{k}m_{\Delta}} \\&\leq \frac{\mathbb{E}_{\tilde{x}_t \sim q_t}\log P_{\phi}(\tilde{x}_t|x_{<t})-\mathbb{E}_{\tilde{x}_t \sim p_t}\log P_{\phi}(\tilde{x}_t|x_{<t})} {(2m_{\phi}+\sqrt{k}m_{\Delta})},
    \end{align*}
Hence,  $\text{TNR}_{\tau}(S_{LAPD})\succeq \text{TNR}_{\tau}(S_{Align}).$

\subsection{Empirical Validation of Assumptions}
\label{app:assumption-validation}

To verify that the distributional relations underlying these assumptions are already visible in finite samples, we extract the logits of all tokens in each benchmark and compute the corresponding token-level averages. Table~\ref{tab:assumption_validation} reports the results.

\Cref{assumption 1} states that the base model is approximately insensitive to the text source in the aggregate. Empirically, the per-token base-model gap $\mathbb{E}_q[\log P_\theta]-\mathbb{E}_p[\log P_\theta]$ ranges from 0.0853 to 0.6446. The aligned-model gap is consistently positive and ranges from 0.3369 to 0.8780, with an average of 0.6248. This average exceeds that of the base model by 50.7\%, supporting the alignment-induced shift formalized by \Cref{assumption 2}.

The variance rows provide a further check of \Cref{assumption 2}. Across all five benchmarks, $\mathrm{Var}_q[\log P_\phi]$ is greater than $\mathrm{Var}_q[\Delta]$, with ratios ranging from 3.9802 to 7.3495. A positive $\epsilon$ is only required for \Cref{theorem 2}. \Cref{theorem 1} also holds with $\epsilon=0$. Finally, \Cref{assumption 3} requires the covariance under human text to be no smaller than that under AI text, and this relation holds on every benchmark. \Cref{assumption 3} is used only in the proof of \Cref{theorem 2}.

These statistics are provided to demonstrate the empirical plausibility of the assumptions rather than as a formal replacement for them.

\begin{table*}[h]
  \caption{Empirical validation of the distributional statistics used in
  Assumptions~1--3. Entries are averages over token-level statistics on each
  benchmark.}
  \label{tab:assumption_validation}
  \begin{center}
  \begin{small}
  \renewcommand{\arraystretch}{1.2}
  \setlength{\tabcolsep}{0pt}
  \begin{tabularx}{\textwidth}{lYYYYY}
    \toprule
    \textbf{Statistic} & \textbf{M4} & \textbf{\makecell[c]{DetectRL \\
    Domain}} & \textbf{\makecell[c]{DetectRL \\ LLM}} & \textbf{RAID} &
    \textbf{RealDet} \\
    \midrule
    \multicolumn{6}{l}{\textit{Base model}} \\
    $\mathbb{E}_q[\log P_\theta]$ & $-1.4206$ & $-1.8007$ & $-1.8478$ &
    $-1.4551$ & $-1.8934$ \\
    $\mathbb{E}_p[\log P_\theta]$ & $-1.8513$ & $-1.8860$ & $-2.2089$ &
    $-2.0065$ & $-2.5380$ \\
    \textbf{Gap ($q-p$)} & \textbf{0.4307} & \textbf{0.0853} &
    \textbf{0.3611} & \textbf{0.5514} & \textbf{0.6446} \\
    \midrule
    \multicolumn{6}{l}{\textit{Aligned model}} \\
    $\mathbb{E}_q[\log P_\phi]$ & $-1.6693$ & $-1.9270$ & $-1.9310$ &
    $-1.5876$ & $-2.0486$ \\
    $\mathbb{E}_p[\log P_\phi]$ & $-2.1796$ & $-2.2639$ & $-2.5737$ &
    $-2.3435$ & $-2.9266$ \\
    \textbf{Gap ($q-p$)} & \textbf{0.5103} & \textbf{0.3369} &
    \textbf{0.6427} & \textbf{0.7559} & \textbf{0.8780} \\
    $\mathrm{Var}_q[\log P_\phi]$ & 6.6297 & 8.2167 & 8.4087 & 6.2520 &
    8.0638 \\
    $\mathrm{Var}_q[\Delta]$ & 1.3169 & 1.1180 & 1.3139 & 1.4552 & 2.0260 \\
    \textbf{Variance ratio} & \textbf{5.0343} & \textbf{7.3495} &
    \textbf{6.3998} & \textbf{4.2963} & \textbf{3.9802} \\
    \midrule
    \multicolumn{6}{l}{\textit{Covariance}} \\
    $\mathrm{Cov}_p(\Delta,\log P_\phi)$ & 2.1782 & 2.5682 & 2.3666 &
    2.2080 & 2.7890 \\
    $\mathrm{Cov}_q(\Delta,\log P_\phi)$ & 2.0641 & 1.7991 & 1.9692 &
    1.8423 & 2.7293 \\
    \textbf{$\mathrm{Cov}_p\geq\mathrm{Cov}_q$} & \textbf{Yes} &
    \textbf{Yes} & \textbf{Yes} & \textbf{Yes} & \textbf{Yes} \\
    \bottomrule
  \end{tabularx}
  \end{small}
  \end{center}
\end{table*}

\section{Details of Experimental Settings}
\label{appendix:exp}
\subsection{Additional Implementation Details}
Consistent with the settings in \citet{zhu2025dna}, we use Falcon-7B-Instruct \citep{penedo2023falcon} as the scoring (reference) model for all  
baselines. For the perturbation (observer) model, Fast-DetectGPT, Lastde++, Binoculars, and DNA-DetectLLM utilize Falcon-7B \citep{penedo2023falcon}, while DetectGPT employs T5-3B \citep{raffel2020t53b} and is restricted to 10 perturbations per sample due to computational overhead.
We set the sample size to 10,000 for both Fast-DetectGPT and LAPD. For Lastde++, we follow the recommended configuration of $s=4, \epsilon=8, \tau^\prime=15$, with 100 generated samples per detection.
During testing, input texts length are truncated to a maximum length of 1024 tokens, while DetectGPT is capped at 512 tokens due to the position limit of T5-3B. 
All experiments are conducted on two NVIDIA RTX 3090 GPUs (24GB VRAM each). 
Default temperature, top-k and other generation parameters are used unless stated otherwise.

\subsection{Details of Used LLMs}
A complete list of the LLMs used in this study is provided in \Cref{tab:model_details}.  All these models are open-sourced and can be downloaded from HuggingFace.
We primarily select models with approximately 7B parameters. This consistency in model scale ensures that the observed performance variations are attributable to the detection methods themselves rather than differences in model capacity.
\vspace{-1em}
\begin{table}[h]
\caption{Details of the LLMs used.}
\label{tab:model_details}
\centering
\begin{small}
\renewcommand{\arraystretch}{1.2} 
\setlength{\tabcolsep}{6pt}       
\begin{tabularx}{\columnwidth}{lXl}
\toprule
\textbf{Model} & \textbf{Model File} & \textbf{Organization}\\
\midrule
Llama-2-7B \citep{touvron2023llama2} & \texttt{meta-llama/Llama-2-7b-hf} & Meta \\
Llama-2-7B-Instruct \citep{touvron2023llama2} & \texttt{meta-llama/Llama-2-7b-chat-hf} & Meta \\
Falcon-7B \citep{penedo2023falcon} & \texttt{tiiuae/falcon-7b} & TII \\
Falcon-7B-Instruct \citep{penedo2023falcon} & \texttt{tiiuae/falcon-7b-instruct} & TII \\
T5-3B \citep{raffel2020t53b} & \texttt{google-t5/t5-3b} & Google AI \\
GPT-J-6B \citep{wang2021gptj6b} & \texttt{EleutherAI/gpt-j-6b} & EleutherAI\\
GPT-J-6B-Instruct \citep{wang2021gptj6b} & \texttt{Provectus/gpt-j-6b-instruct} & EleutherAI \\
Llama-3.1-8B \citep{grattafiori2024llama3} & \texttt{meta-llama/Llama-3.1-8B} & Meta \\
Llama-3.1-8B-Instruct \citep{grattafiori2024llama3} & \texttt{meta-llama/Llama-3.1-8B-Instruct} & Meta \\
\bottomrule
\end{tabularx}
\end{small}
\end{table}

\subsection{Details of Benchmarks}

The following sections give a brief introduction about the characteristics of the four benchmarks we used in this study. 
The detailed domains and source LLMs each benchmark contains are listed in \Cref{tab:appendix_benchmark_details}.

\textbf{M4} \citep{wang2024m4}. The Multi-generator, Multi-domain, and Multi-lingual dataset is a large-scale benchmark for AI-generated text detection in black-box scenarios. It contains human and AI texts across 7 domains and 7 languages. AI texts are generated by 6 model families, including GPT-4, ChatGPT and BLOOMz.

\textbf{DetectRL} \citep{wu2024detectrl}. This benchmark dataset evaluates the reliability of detectors in realistic scenarios by simulating human revisions, prompts, and writing noises. The \textbf{Multi-LLM} setting uses generations from models like GPT-3.5-Turbo, Claude-Instant, and Llama-2-70B to assess the model-agnosticism, while \textbf{Multi-Domain} spans academic abstracts, news, creative stories, and social reviews.

\textbf{RAID} \citep{dugan2024raid}. This is a large-scale benchmark for assessing the robustness of detectors across 11 models and 8 domains. It contains 11 types of adversarial attacks, along with 4 different decoding strategies, providing a rigorous environment to test how well detection methods handle realistic modifications.

\textbf{RealDet} \citep{zhu2025realdet}. RealDet is a high-reliability bilingual benchmark that emphasizes realistic detection settings and threshold calibration. It spans 15 distinct textual domains and 22 popular LLMs, notably including recent reasoning models such as DeepSeek-R1.

\begin{table*}[h]
\caption{The comprehensive landscape of evaluation benchmarks, detailing the domains and generative source models of each benchmark.}
\label{tab:appendix_benchmark_details}
\centering
\begin{small}
\renewcommand{\arraystretch}{1.2} 
\begin{tabularx}{\textwidth}{lXX}
\toprule
\textbf{Benchmarks} & \textbf{Domains} & \textbf{Source LLMs} \\
\midrule
\textbf{M4} & Wikipedia, WikiHow, Reddit, arXiv, PeerRead, Baike/Web QA, News, Arabic Wikipedia. & GPT-4, ChatGPT, GPT-3.5, Cohere, Dolly-v2, BLOOMz-176B. \\ \midrule
\textbf{DetectRL} & Academic Abstracts, News Articles, Creative Stories, Social Reviews. & GPT-3.5-turbo, PaLM-2-bison, Claude-instant, Llama-2-70b. \\ \midrule
\textbf{RAID} & Abstracts, Books, News, Poetry, Recipes, Reddit, Reviews, Wikipedia, Python Code, Czech News, German News. & GPT-2 XL, GPT-3, ChatGPT, GPT-4, Cohere, Mistral-7B, MPT-30B, LLaMA 2 70B. \\ \midrule
\textbf{RealDet} & ELI5, WikiQA, Wikipedia, Medical Dialog, FiQA, XSum, TLDR, BBC News, WritingPrompt, ROC Stories, Yelp, IMDB, CMV, Abstracts, SQuAD, WebTextQA, BaikeQA, BaiduBaike, NLPCC-DBQA, FinanceZhidao, Chinese Psychological QA, LegalQA. & DeepSeek-R1, GPT-4, ChatGPT, PaLM 2, Ernie Bot, Spark Desk, Qwen turbo, 360GPT S2 V9, Minimax abab 5.5, LLaMA2-13B, ChatGLM2-6B, MOSS-moon-003, MPT-7B, InternLM-7B, Alpaca-7B, Guanaco-7B, Vicuna-13B, BLOOMz-7B, Falcon-7B, OPT-6.7B, Baichuan-13B, Flan-T5-XXL. \\
\bottomrule
\end{tabularx}
\end{small}
\end{table*}

\section{Additional Results}
\subsection{Details of TPR at low FPR}
\label{appendix:tpr_results}

The complete results of TPR at $ 0.5\% $ FPR across different domains are presented in Table \ref{tab:tpr_results}. 

Existing methods generally exhibit poor performance under this strict metric: Fast-DetectGPT, Binoculars, and DNA-DetectLLM achieve average TPRs of $ 51.79\% $, $ 66.67\% $, and $ 59.56\% $, respectively. 

On the contrary, RAI demonstrates performance comparable to the best baseline, while LAPD attains an average TPR of $ 92.27\% $, obtaining a substantial relative improvement of $ 76.81\% $.
Notably, on the XSum dataset, most baselines suffer significant performance degradation. However, both RAI and LAPD maintain robust detection abilities, consistently outperforming other methods.

\begin{table*}[h]
\caption{Detailed detection results of TPR (\%) at $0.5\%$ FPR across
three datasets. Source models are abbreviated as: \textbf{GPT}
(GPT-4 Turbo), \textbf{Gemini} (Gemini-2.0 Flash), and
\textbf{Claude} (Claude-3.7 Sonnet). The best results are highlighted
in \textbf{bold}. Methods with clubs ($\clubsuit$) standardize
the scores by perturbing or generating auxiliary sequences.}
\label{tab:tpr_results}
\begin{center}
\begin{small}
\renewcommand{\arraystretch}{1.2}
\setlength{\tabcolsep}{0pt} 
\begin{tabularx}{\textwidth}{lYYYYYYYYYY} 
\toprule
\multirow{2}{*}{\textbf{Methods}} & \multicolumn{3}{c}{\textbf{XSum}} & \multicolumn{3}{c}{\textbf{WritingPrompts}} & \multicolumn{3}{c}{\textbf{Arxiv}} & \multirow{2}{*}{\textbf{Avg.}} \\
\cmidrule(lr){2-4} \cmidrule(lr){5-7} \cmidrule(lr){8-10}
 & GPT & Gemini & Claude & GPT & Gemini & Claude & GPT & Gemini & Claude & \\
    \midrule
    Entropy& 0.30  & 3.90  & 0.30  & 0.10  & 0.20  & 0.00  & 0.60  & 0.00  & 0.10  & 0.61  \\
    Likelihood           & 0.00  & 0.10  & 0.00  & 8.80  & 51.40 & 16.20 & 1.90  & 27.10 & 7.00  & 12.50 \\
    LogRank              & 0.10  & 0.70  & 0.10  & 10.40 & 50.00 & 16.80 & 2.40  & 31.90 & 7.20  & 13.29 \\
    DetectGPT$^\clubsuit$      & 0.30  & 0.30  & 0.30  & 0.70  & 1.00  & 0.40  & 0.00  & 0.30  & 0.10  & 0.38  \\
    Fast-DetectGPT$^\clubsuit$ & 63.50 & 57.60 & 27.70 & 41.10 & 87.70 & 16.40 & 23.20 & 90.40 & 58.50 & 51.79 \\
    Lastde++$^\clubsuit$       & 51.70 & 51.20 & 19.00 & 22.00 & 78.70 & 5.20  & 23.10 & 85.90 & 54.90 & 43.52 \\
    Binoculars           & 57.30 & 77.60 & 56.30 & 52.30 & 97.40 & 66.20 & 42.70 & 94.30 & 55.90 & 66.67 \\
    DNA-DetectLLM$^\clubsuit$  & 76.00 & 1.30  & 48.30 & 64.80 & 85.60 & 56.40 & 32.10 & 96.90 & 74.60 & 59.56 \\
    \midrule
    RAI                  & 86.50 & 76.70 & 69.90 & 62.40 & 79.90 & 50.90 & 57.60 & 81.80 & 63.60 & 69.92 \\
    LAPD$^\clubsuit$     & \textbf{97.90} & \textbf{85.00} & \textbf{97.20}
                         & \textbf{92.60} & \textbf{94.70} & \textbf{95.40}
                         & \textbf{68.20} & \textbf{99.60} & \textbf{99.80}
                         & \textbf{92.27} \\
    \bottomrule
\end{tabularx}
\end{small}
\end{center}
\vspace{-1em}
\end{table*}

\subsection{ROC Curves on various models and datasets}

\begin{figure*}[h]
  \centering
  \hspace{0.03\textwidth}
  \begin{minipage}{0.31\textwidth}\centering XSum \end{minipage} \hfill
  \begin{minipage}{0.31\textwidth}\centering WritingPrompts \end{minipage} \hfill
  \begin{minipage}{0.31\textwidth}\centering Arxiv \end{minipage}
  
  \vspace{0.5em}

  \begin{minipage}[c]{0.03\textwidth}
    \rotatebox{90}{\small GPT-4-Turbo}
  \end{minipage}
  \begin{minipage}[c]{0.95\textwidth}
    \begin{subfigure}{0.31\linewidth}
      \includegraphics[width=\linewidth]{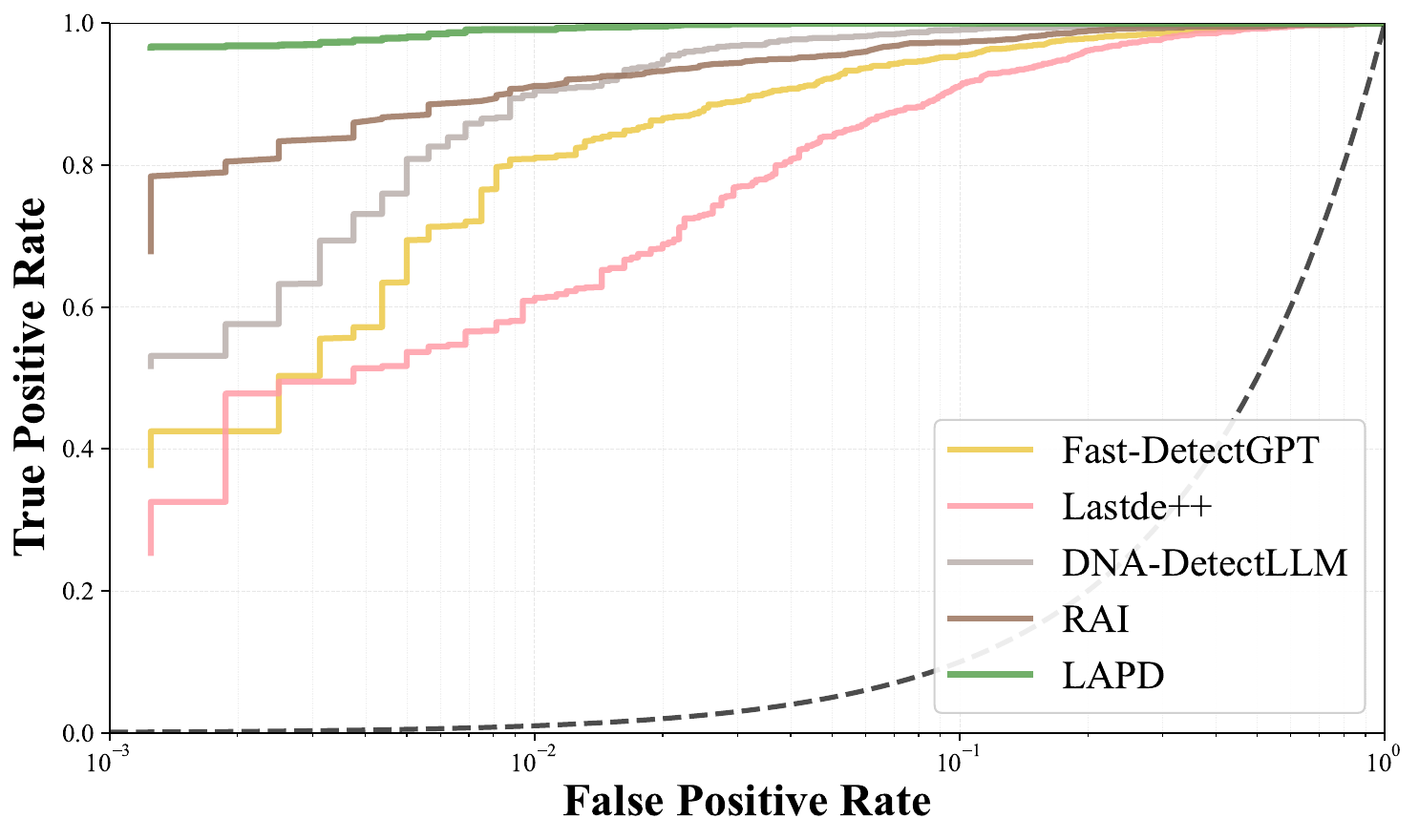}
    \end{subfigure}\hfill
    \begin{subfigure}{0.31\linewidth}
      \includegraphics[width=\linewidth]{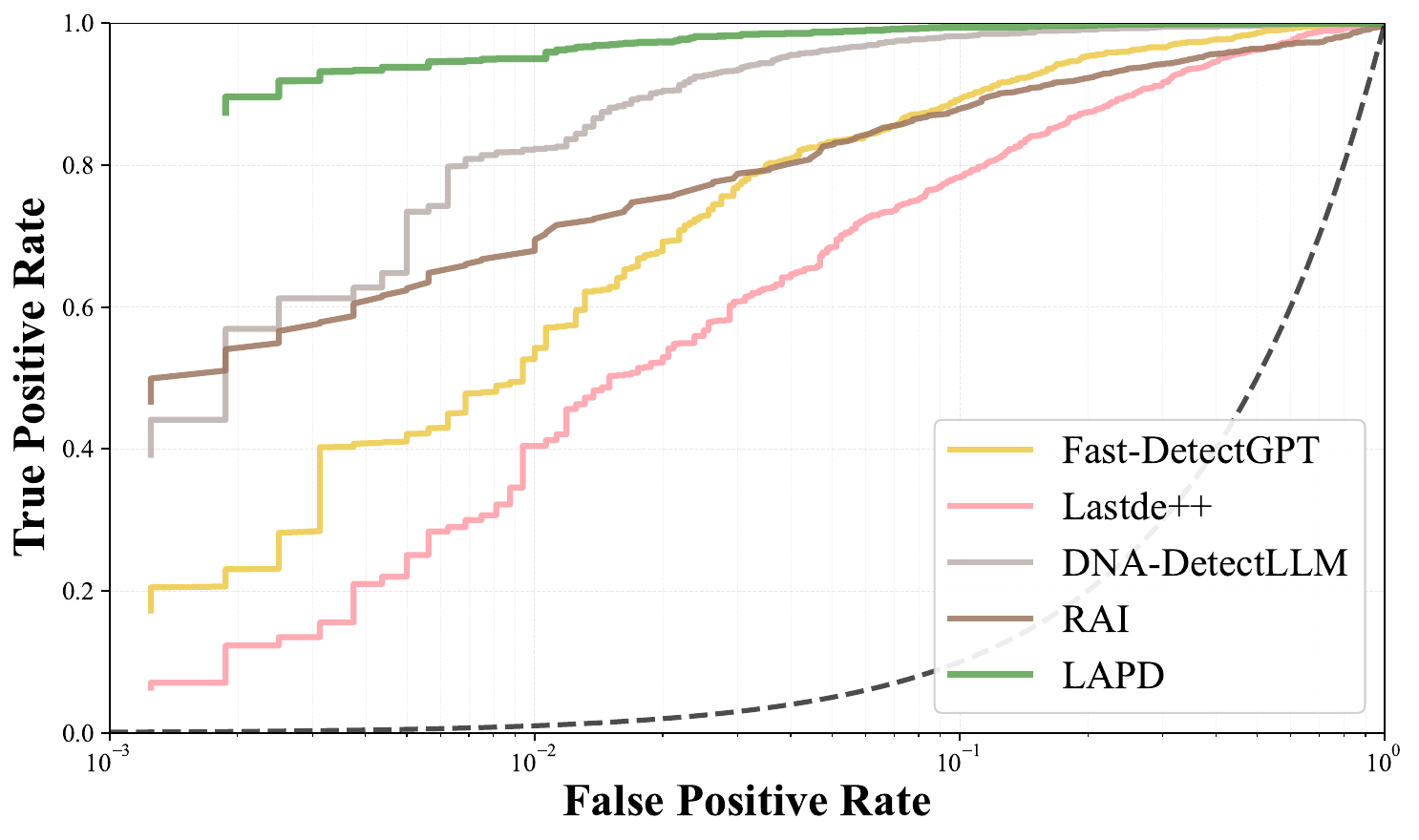}
    \end{subfigure}\hfill
    \begin{subfigure}{0.31\linewidth}
      \includegraphics[width=\linewidth]{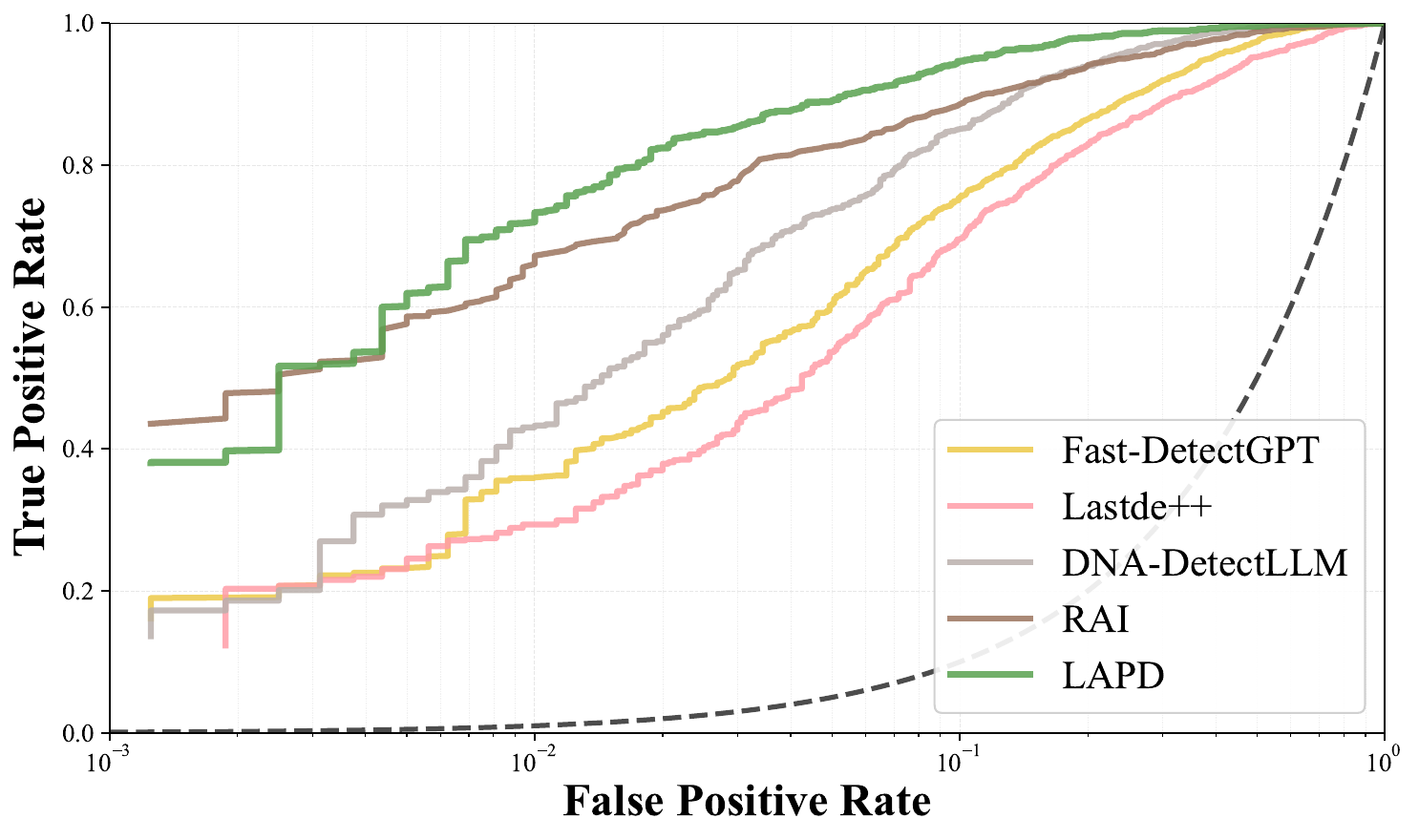}
    \end{subfigure}
  \end{minipage}

  \vspace{0.5em}

  \begin{minipage}[c]{0.03\textwidth}
    \rotatebox{90}{\small Gemini-2.0-Flash}
  \end{minipage}
  \begin{minipage}[c]{0.95\textwidth}
    \begin{subfigure}{0.31\linewidth}
      \includegraphics[width=\linewidth]{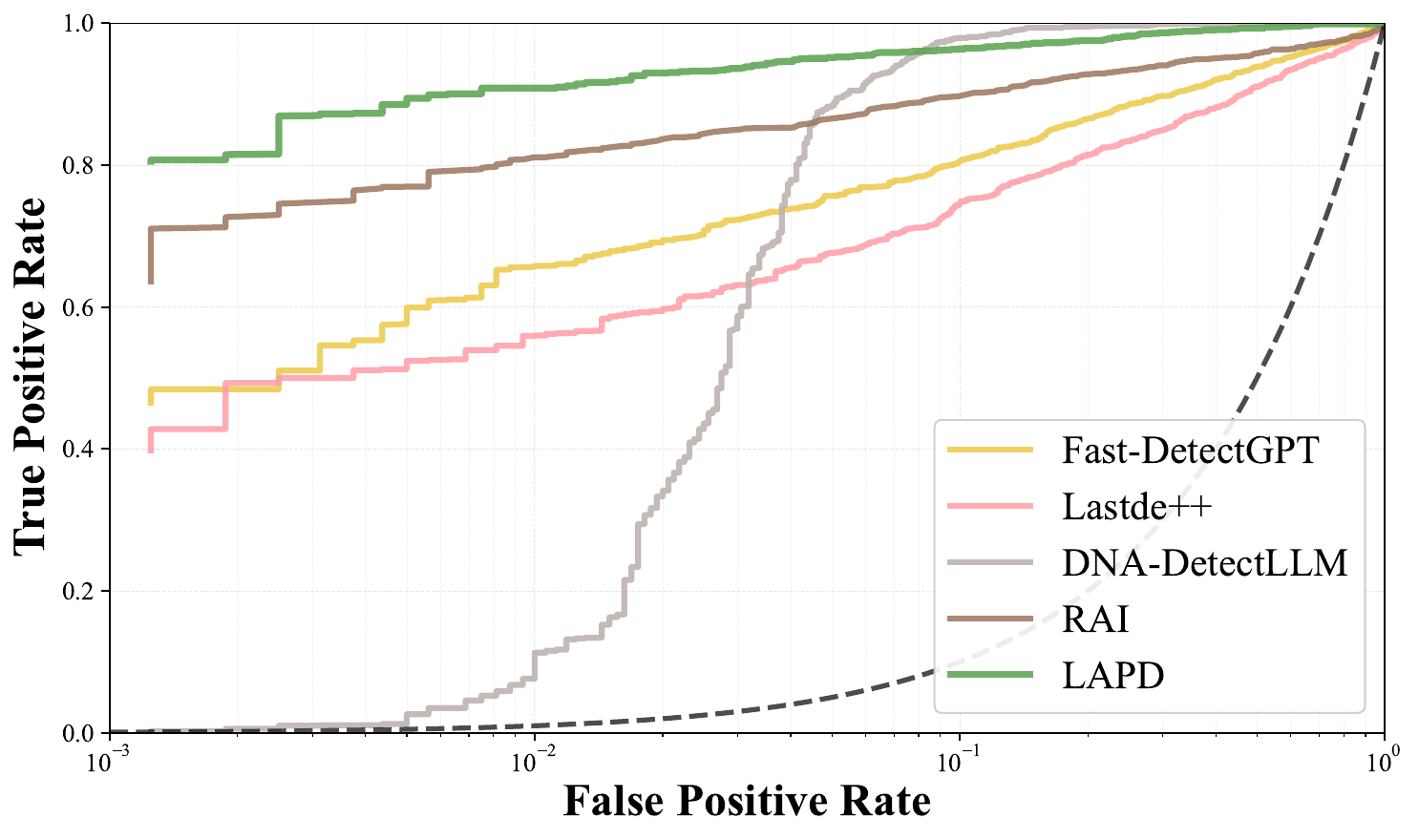}
    \end{subfigure}\hfill
    \begin{subfigure}{0.31\linewidth}
      \includegraphics[width=\linewidth]{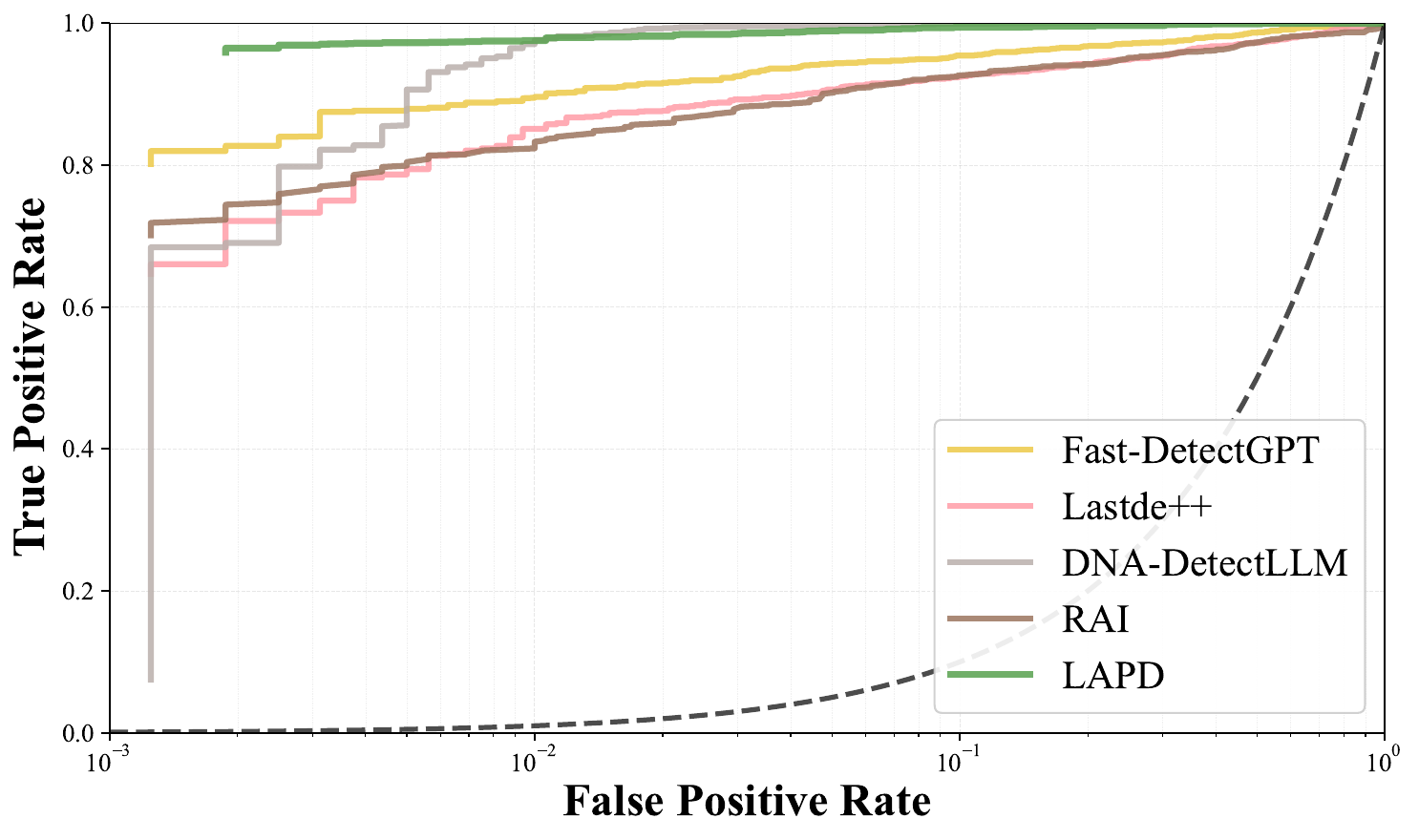}
    \end{subfigure}\hfill
    \begin{subfigure}{0.31\linewidth}
      \includegraphics[width=\linewidth]{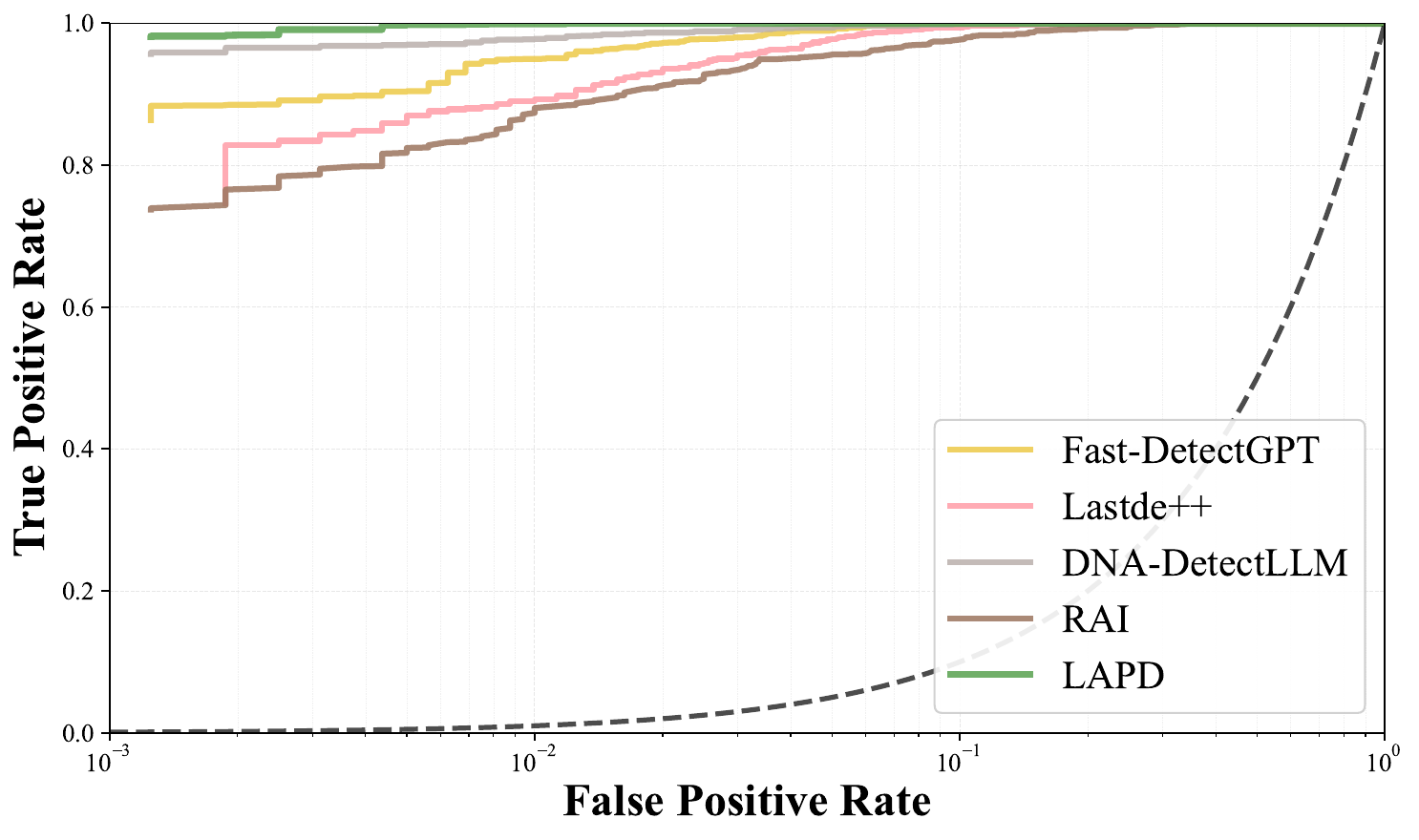}
    \end{subfigure}
  \end{minipage}

  \vspace{0.5em}

  \begin{minipage}[c]{0.03\textwidth}
    \rotatebox{90}{\small Claude-3.7-Sonnet}
  \end{minipage}
  \begin{minipage}[c]{0.95\textwidth}
    \begin{subfigure}{0.31\linewidth}
      \includegraphics[width=\linewidth]{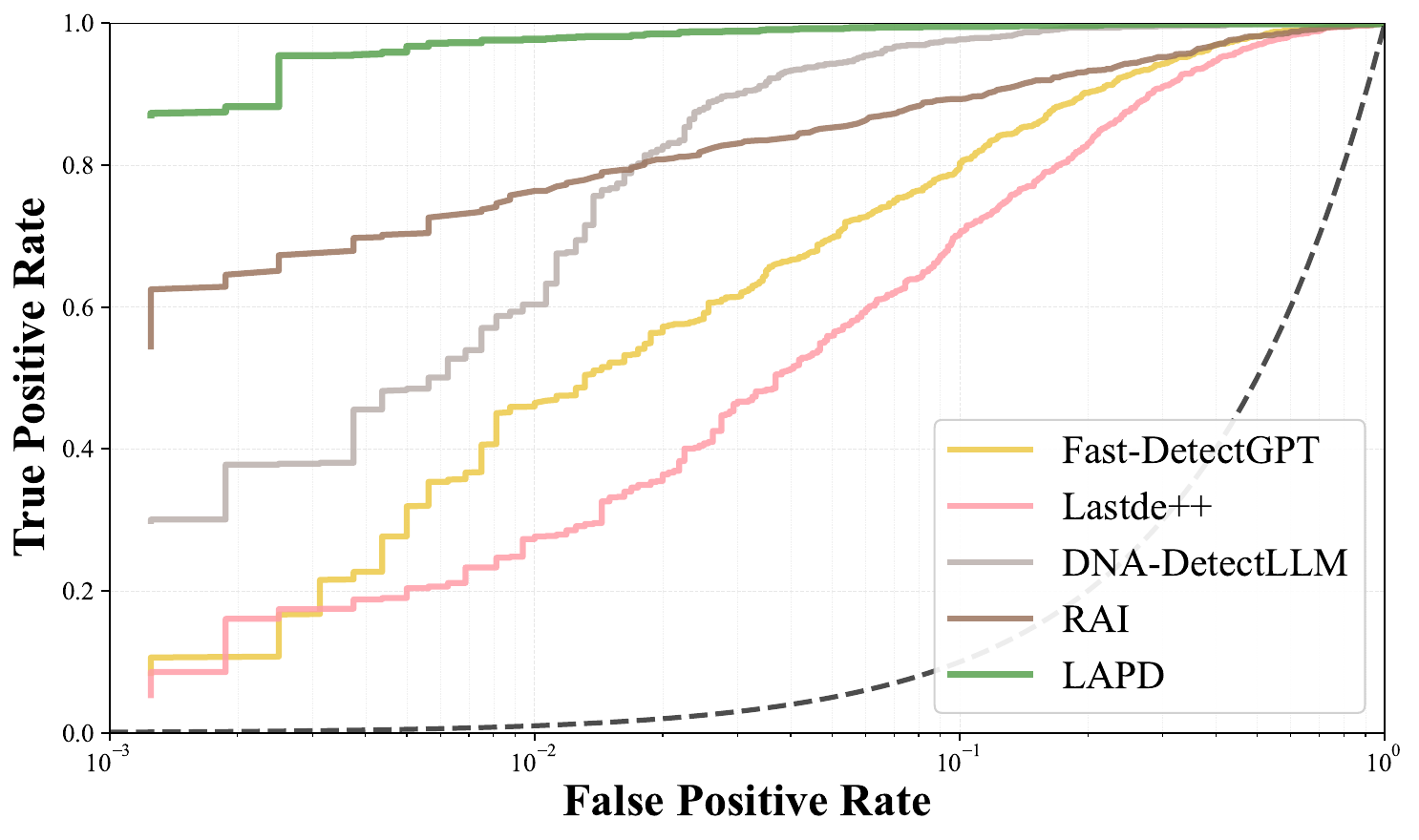}
    \end{subfigure}\hfill
    \begin{subfigure}{0.31\linewidth}
      \includegraphics[width=\linewidth]{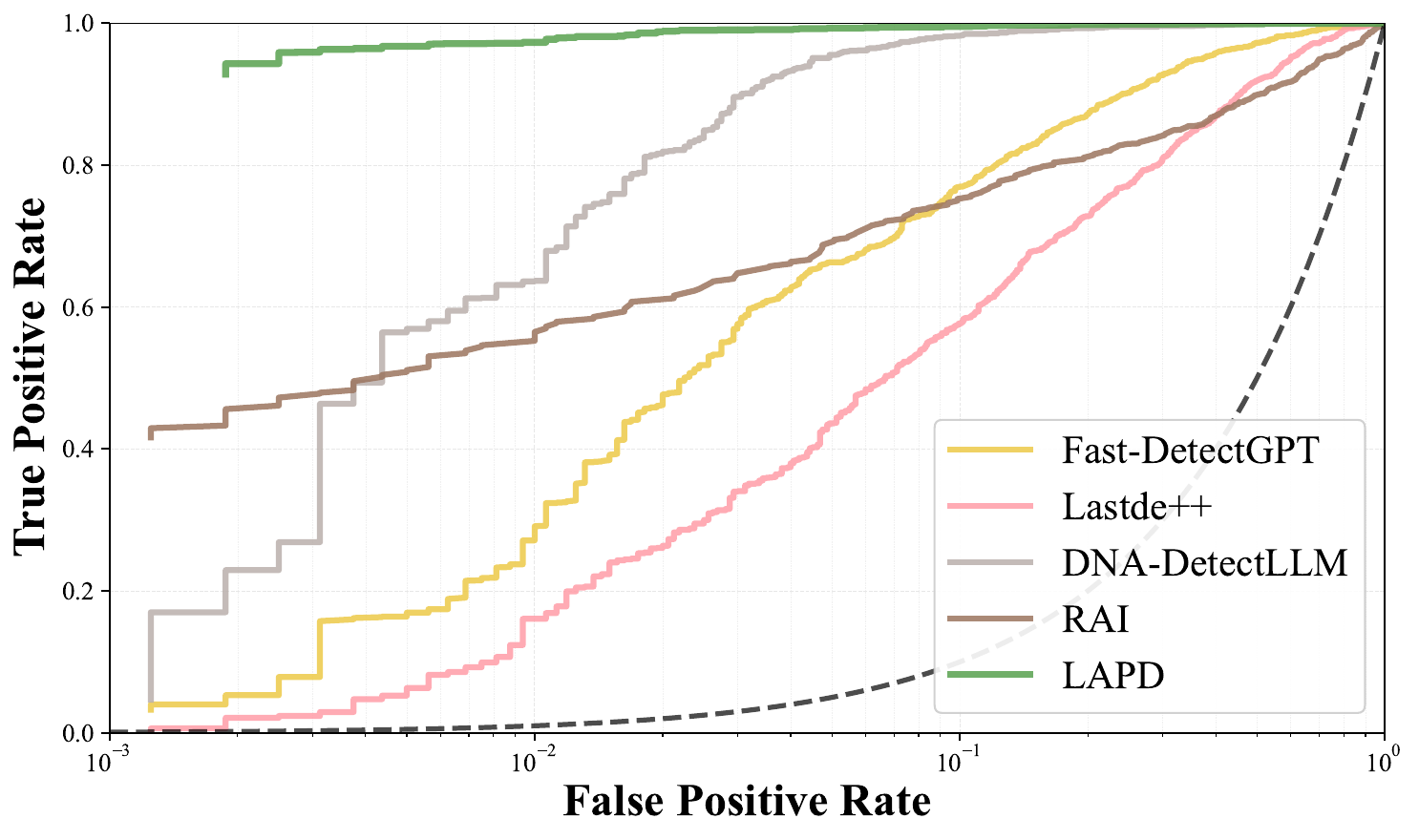}
    \end{subfigure}\hfill
    \begin{subfigure}{0.31\linewidth}
      \includegraphics[width=\linewidth]{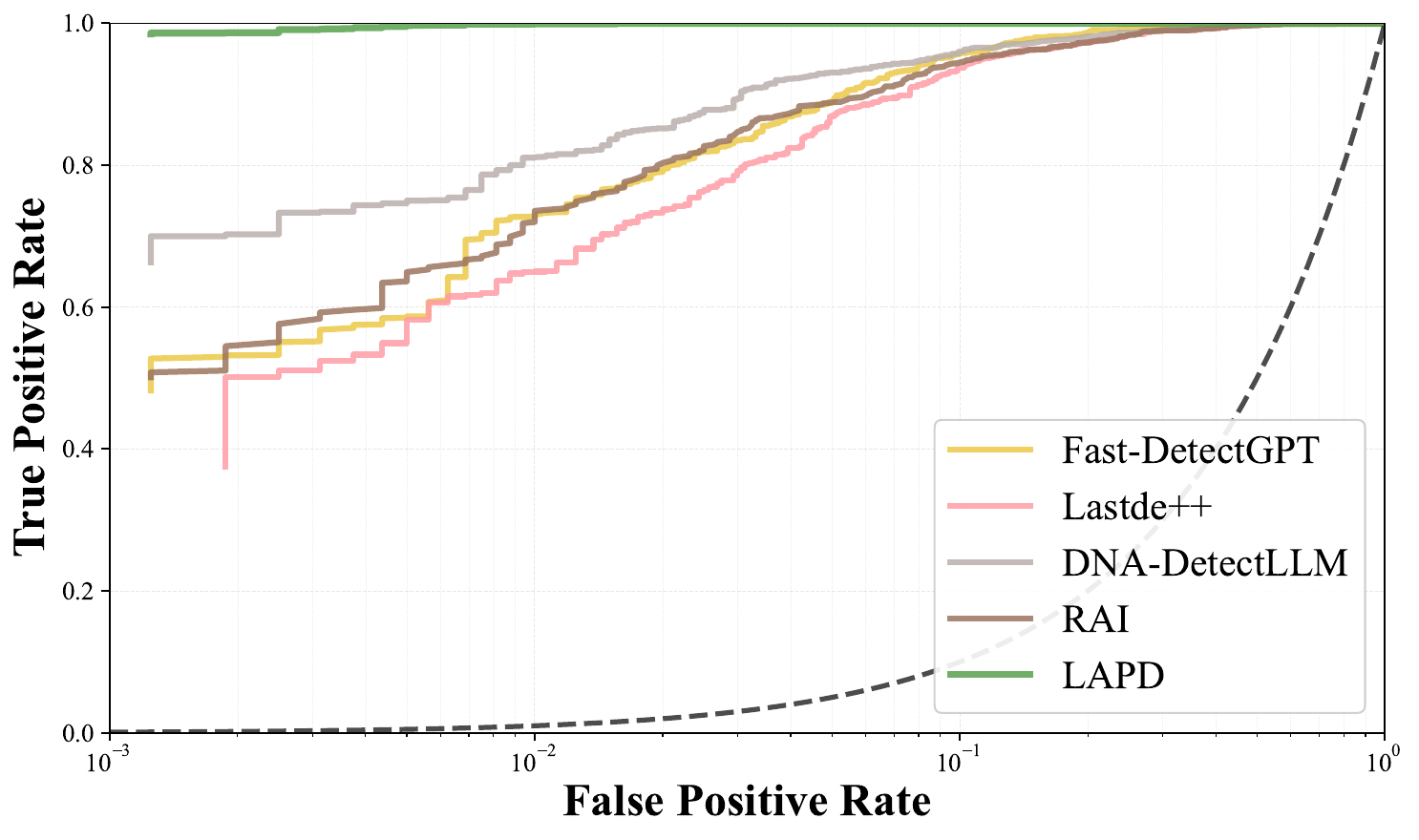}
    \end{subfigure}
  \end{minipage}

  \caption{Log-scale ROC curves evaluated across three datasets (columns) and three source models (rows). The dashed lines denote the random classifiers.}
  \label{fig:roc_curves_matrix}
\end{figure*}

\Cref{fig:roc_curves_matrix} presents the log-scale ROC curves of five representative methods across three datasets (columns) and three source models (rows). 
To emphasize the high-confidence detection performance, the FPR is visualized on a logarithmic scale starting from $ 0.1\% $ to $100\%$.

Across all experimental configurations, RAI shows competitive results, while LAPD consistently maintains a significantly higher TPR compared to all the baselines, particularly in the minimal FPR from $ 0.1\% $ to 1\%. 
These observations validate the cross-domain and cross-model robustness of LAPD in practical scenarios requiring minimal FPRs. 

\subsection{Cross-Domain Threshold Transfer}
\label{app:cross-domain-threshold-transfer}

The experiments above calibrate the detection threshold within each dataset. In practical deployments, repeating this calibration for every target domain may not be feasible. We therefore test whether a single threshold can be transferred across domains. Specifically, for each detector, we determine the threshold that yields an FPR of 0.5\% on XSum and keep this threshold fixed when evaluating WritingPrompts and Arxiv.

Table~\ref{tab:cross_domain_threshold_transfer} reports the resulting FPR and TPR. LAPD is the only method that transfers a single threshold to the other domains while maintaining an FPR below 2\% and a TPR above 90\%. These results demonstrate the robustness of LAPD to threshold selection and support its practical viability in real-world deployment.

\begin{table*}[h]
  \caption{Cross-domain threshold transfer results (\%). For each method, the threshold is calibrated on XSum to achieve an FPR of 0.5\% and is then kept fixed on WritingPrompts and Arxiv.}
  \label{tab:cross_domain_threshold_transfer}
  \begin{center}
  \begin{small}
  \renewcommand{\arraystretch}{1.2}
  \setlength{\tabcolsep}{0pt}
  \begin{tabularx}{\textwidth}{lYYYYYYY}
    \toprule
    \multirow{2}{*}{\textbf{Methods}} & \multirow{2}{*}{\textbf{Threshold}} &
    \multicolumn{2}{c}{\textbf{XSum}} &
    \multicolumn{2}{c}{\textbf{WritingPrompts}} &
    \multicolumn{2}{c}{\textbf{Arxiv}} \\
    \cmidrule(lr){3-4}
    \cmidrule(lr){5-6}
    \cmidrule(lr){7-8}
     & & FPR & TPR & FPR & TPR & FPR & TPR \\
    \midrule
    Fast-DetectGPT & 3.0593 & 0.5 & 53.8 & 1.5 & 64.0 & 0.0 & 39.8 \\
    Binoculars & $-0.8594$ & 0.5 & 64.2 & 0.4 & 65.5 & 0.3 & 61.8 \\
    DNA-DetectLLM & $-0.5119$ & 0.5 & 67.8 & 0.0 & 72.0 & 4.3 & 92.7 \\
    LAPD & 2.3692 & 0.5 & 93.3 & \textbf{0.6} & \textbf{96.3} & \textbf{1.2} & \textbf{91.5} \\
    \bottomrule
  \end{tabularx}
  \end{small}
  \end{center}
\end{table*}

\section{Experimental Results under Attacks}
\label{appendix:realistic_attacks}

\Cref{tab:attack_results} illustrates the performance under attacks. We apply three types of attacks: insertion, deletion, and replacement to 1\% of the tokens in AI-generated texts from GPT-4 Turbo, Gemini-2.0 Flash, and Claude-3.7 Sonnet.
Experimental results demonstrate that while the baselines suffer from varying degrees of performance degradation, LAPD exhibits negligible performance drop.

\begin{table*}[t]
\caption{Robustness against various editing attacks (AUROC \%). We evaluate three types of random editing: Insert, Delete, and Replace at a 1\% token ratio. The best results are highlighted in \textbf{bold}. Methods with clubs ($\clubsuit$) standardize the scores by perturbing or generating auxiliary sequences.}
\vspace{-0.5em}
\label{tab:attack_results}
\begin{center}
\begin{small}
\renewcommand{\arraystretch}{1.2}
\setlength{\tabcolsep}{0pt}
\begin{tabularx}{\textwidth}{lYYYYYYYYY}
\toprule
\multirow{2}{*}{\textbf{Methods}} & \multicolumn{3}{c}{\textbf{GPT-4 Turbo}} & \multicolumn{3}{c}{\textbf{Gemini-2.0 Flash}} & \multicolumn{3}{c}{\textbf{Claude-3.7 Sonnet}} \\
\cmidrule(lr){2-4} \cmidrule(lr){5-7} \cmidrule(lr){8-10}
 & \small Insert & \small Delete & \small Replace & \small Insert & \small Delete & \small Replace & \small Insert & \small Delete & \small Replace \\
\midrule
Entropy & 57.66 & 63.51 & 55.59 & 65.70 & 71.20 & 63.58 & 73.42 & 78.36 & 70.98 \\
Likelihood & 51.54 & 60.73 & 49.91 & 76.39 & 82.85 & 74.74 & 64.13 & 73.03 & 61.94 \\
LogRank & 52.21 & 60.69 & 50.47 & 76.79 & 82.95 & 74.96 & 64.24 & 72.78 & 61.99 \\
DetectGPT\perturb & 45.71 & 46.65 & 46.15 & 47.77 & 51.63 & 47.43 & 47.18 & 49.83 & 46.16 \\
Fast-DetectGPT\perturb & 84.82 & 90.85 & 84.57 & 94.37 & 95.82 & 94.16 & 84.32 & 91.29 & 83.23 \\
Lastde++\perturb & 68.89 & 85.13 & 68.36 & 86.36 & 92.51 & 85.88 & 66.78 & 84.77 & 66.09 \\
Binoculars & 87.69 & 93.43 & 87.46 & 97.43 & 98.07 & 97.37 & 89.93 & 95.41 & 89.00 \\
DNA-DetectLLM\perturb & 93.48 & 96.34 & 93.16 & 97.73 & 98.17 & 97.70 & 93.84 & 96.93 & 93.32 \\
\midrule
RAI & 92.83 & 91.81 & 86.92 & 95.45 & 94.58 & 95.54 & 92.99 & 91.44 & 93.60 \\
LAPD\perturb & \textbf{99.16} & \textbf{98.99} & \textbf{99.05} & \textbf{99.34} & \textbf{99.37} & \textbf{99.36} & \textbf{99.64} & \textbf{99.68} & \textbf{99.70} \\
\bottomrule
\end{tabularx}
\end{small}
\end{center}
\end{table*}

Currently, there are many different attack methods \cite{fang2025your}. So we further evaluated the robustness of the methods against more realistic attacks from the RAID benchmark. 
Specifically, we consider four representative attack strategies: semantic-level attacks (paraphrase and synonym substitution) and character-level attacks (homoglyph and zero-width space insertion), together with a no-attack control condition. 
For each type of attack, we construct a balanced evaluation set of 1,000 samples.

As shown in \Cref{tab:appendix_realistic_attacks}, our LAPD method outperforms all baselines, achieving the highest average attack AUROC of 83.36\% and the highest clean performance of 89.96\%. This experiment further strengthens the rigor and fairness of our evaluation.

\begin{table*}[h]
\caption{Detection performance (AUROC \%) under 4 representative attack types. The best results are highlighted in \textbf{bold}. Methods with clubs ($\clubsuit$) standardize the scores by perturbing or generating auxiliary sequences.}
\label{tab:appendix_realistic_attacks}
\begin{center}
\begin{small}
\renewcommand{\arraystretch}{1.2}
\setlength{\tabcolsep}{0pt} 
\begin{tabularx}{\textwidth}{lYYYYYY}
\toprule
\textbf{Method} & \textbf{Homoglyph} & \textbf{Paraphrase} & \textbf{Synonym} & \textbf{\makecell[c]{Zero-width \\ Space}} & \textbf{\makecell[c]{Attack \\ Avg.}} & \textbf{None} \\
\midrule
Fast-DetectGPT\perturb & 67.21 & 72.82 & 80.49 & 63.36 & 70.97 & 84.95 \\
Lastde++\perturb & 65.25 & 72.65 & 80.15 & 61.73 & 69.95 & 89.20 \\
Binoculars & 71.45 & 80.25 & 80.70 & 64.03 & 74.11 & 85.59 \\
DNA-DetectLLM\perturb & 68.29 & 81.10 & 82.99 & 59.83 & 73.05 & 87.32 \\
\midrule
RAI & 48.09 & 68.90 & 68.50 & 36.93 & 55.61 & 78.06 \\
LAPD\perturb & \textbf{82.71} & \textbf{82.53} & \textbf{87.09} & \textbf{81.10} & \textbf{83.36} & \textbf{89.96} \\
\bottomrule
\end{tabularx}
\end{small}
\end{center}
\end{table*}

\section{Experimental Results under Cross-family Model Pairs}

To analyze the performance of the detectors when the aligned model comes from a different family from the base model, we conducted an ablation study using cross-family model pairs. 
Note that the methods using a sampling model for perturbation-based standardization such as Fast-DetectGPT, Lastde++, and LAPD, require the sampling model and the scoring model to share a same tokenizer, which prevents them from being applied in this cross-model setting. We therefore report only the results of RAI, which does not rely on a sampling model.

\Cref{tab:appendix_cross_family} demonstrates that using a cross-family pair leads to an average AUROC drop ranging from 8.73\% to 29.80\%. 
This validates our derivation that the aligned model should come from the base model by the alignment process, so that the detection statistic $\Delta(\boldsymbol{x})$ captures the alignment imprint without interference from pretraining distribution shifts.

\vspace{1em}

\begin{table*}[h]
\caption{Detection performance (AUROC \%) when the base and instruct models come from different model families. The last row corresponds to the original RAI configuration with a matched Llama2-7B pair. The best results are highlighted in \textbf{bold}.}
\label{tab:appendix_cross_family}
\begin{center}
\begin{small}
\renewcommand{\arraystretch}{1.2}
\begin{tabularx}{\textwidth}{l
>{\centering\arraybackslash\hsize=0.8\hsize}X
>{\centering\arraybackslash\hsize=1.4\hsize}X
>{\centering\arraybackslash\hsize=1.4\hsize}X
>{\centering\arraybackslash\hsize=0.8\hsize}X
>{\centering\arraybackslash\hsize=0.8\hsize}X
>{\centering\arraybackslash\hsize=0.8\hsize}X
}
\toprule
\textbf{Base/Instruct Model} &
\textbf{M4} &
\textbf{\makecell[c]{DetectRL \\ LLM}} &
\textbf{\makecell[c]{DetectRL \\ Domain}} &
\textbf{RAID} &
\textbf{RealDet} &
\textbf{Avg.} \\
\midrule
Falcon-7B/Llama2-7B-Instruct   & 55.87 & 90.27 & 92.35 & 67.22 & 78.61 & 76.86 \\
GPT-J-6B/Llama2-7B-Instruct    & 55.22 & 84.41 & 89.38 & 66.85 & 67.87 & 72.75 \\
Llama2-7B/Falcon-7B-Instruct   & 63.33 & 53.65 & 45.33 & 64.69 & 65.95 & 58.59 \\
Llama2-7B/GPT-J-6B-Instruct    & 53.36 & 49.66 & 43.49 & 60.23 & 72.24 & 55.80 \\
\midrule
Llama2-7B/Llama2-7B-Instruct   & \textbf{70.46} & \textbf{93.13} & \textbf{96.19} & \textbf{81.28} & \textbf{86.92} & \textbf{85.60} \\
\bottomrule
\end{tabularx}
\end{small}
\end{center}
\end{table*}

\vspace{1em}

\section{Experimental Results under Same Proxy Model}

In the main experiments, we follow the official implementations of the baselines, which use Falcon-7B-Instruct as the scoring model, while LAPD uses the Llama-2-7B model pair for detection.
To eliminate the potential confounding factor introduced by the proxy model, we evaluated the baselines using the Llama-2-7B model pair.
As summarized in \Cref{tab:appendix_matched_comparison}, LAPD still consistently achieves SOTA performance across all benchmarks, confirming that the performance gains stem from the method itself rather than from the choice of scoring model.

\vspace{1em}

\begin{table*}[h]
\caption{Matched comparison of detection performance (AUROC \%) across baselines and LAPD. All methods use the Llama2-7B base/instruct pair for detection. The best results are highlighted in \textbf{bold}. Methods with clubs ($\clubsuit$) standardize the scores by perturbing or generating auxiliary sequences.}
\label{tab:appendix_matched_comparison}
\begin{center}
\begin{small}
\renewcommand{\arraystretch}{1.2}
\begin{tabularx}{\textwidth}{lYYYYYY}
\toprule
\textbf{Methods} & \textbf{M4} & \textbf{\makecell[c]{DetectRL \\ LLM}} & \textbf{\makecell[c]{DetectRL \\ Domain}} & \textbf{RAID} & \textbf{RealDet} & \textbf{Avg.} \\
\midrule
Fast-DetectGPT\perturb & 86.14 & 92.42 & 86.42 & 84.77 & 94.16 & 88.78 \\
Lastde++\perturb       & 86.69 & 86.59 & 78.57 & 86.88 & 94.02 & 86.55 \\
Binoculars             & 87.27 & 93.11 & 88.33 & 85.30 & 94.56 & 89.72 \\
DNA-DetectLLM\perturb  & 85.79 & 92.98 & 87.72 & 84.55 & 94.63 & 89.13 \\
\midrule
RAI                    & 70.46 & 93.13 & \textbf{96.19} & 81.28 & 86.92 & 85.60 \\
LAPD\perturb           & \textbf{88.02} & \textbf{97.17} & 96.11 & \textbf{85.21} & \textbf{95.32} & \textbf{92.37} \\
\bottomrule
\end{tabularx}
\end{small}
\end{center}
\end{table*}

\vspace{1em}

\section{Comparison with supervised methods}

We also compare LAPD with supervised (or hybrid) detection methods. Specifically, we tested three representative baselines, OpenAI Detector \cite{solaiman2019release}, ReMoDetect \cite{lee2024remodetect}, and ImBD \cite{chen2025imitate}, and evaluated in four benchmarks. 

As demonstrated in \Cref{tab:appendix_supervised_comparison}, LAPD achieves the best average AUROC of 92.37\%, outperforming all the baselines. This shows that LAPD maintains a significant advantage over strong supervised and hybrid detection methods.

\begin{table*}[h]
\caption{Detection performance (AUROC, \%) of LAPD compared with representative supervised and hybrid baselines. The best results are highlighted in 	\textbf{bold}.}
\label{tab:appendix_supervised_comparison}
\begin{center}
\begin{small}

\renewcommand{\arraystretch}{1.2}
\begin{tabularx}{\textwidth}{lYYYYYY}
\toprule
\textbf{Methods} & \textbf{M4} & \textbf{\makecell[c]{DetectRL \\ LLM}} & \textbf{\makecell[c]{DetectRL \\ Domain}} & \textbf{RAID} & \textbf{RealDet} & \textbf{Avg.} \\
\midrule
OpenAI Detector & 72.24 & 78.15 & 74.60 & 73.37 & 84.96 & 76.66 \\
ReMoDetect  & 74.98 & 89.11 & 80.93 & 66.89 & 92.18 & 80.82 \\
ImBD        & 73.41 & 83.25 & 89.81 & 77.77 & 92.12 & 83.27 \\
\midrule
LAPD        & 	\textbf{88.02} & 	\textbf{97.17} & 	\textbf{96.11} & 	\textbf{85.21} & 	\textbf{95.32} & 	\textbf{92.37} \\
\bottomrule
\end{tabularx}
\end{small}
\end{center}
\end{table*}

\end{document}